\documentclass[lettersize, journal, 10pt, twoside]{IEEEtran}
\usepackage{amsmath, amssymb, amsthm}
\usepackage{bm}
\usepackage{mathrsfs}
\usepackage[ruled, linesnumbered]{algorithm2e}
\usepackage[caption=false,font=normalsize,labelfont=sf,textfont=sf]{subfig}
\usepackage{textcomp}
\usepackage{stfloats}
\usepackage{url}
\usepackage{verbatim}
\usepackage{graphicx}
\usepackage{cite}
\usepackage{multirow}
\usepackage{multicol}
\usepackage{makecell}
\usepackage{booktabs}
\usepackage{xcolor}
\usepackage{gensymb}
\usepackage{hyperref}
\usepackage{footnote}
\usepackage{tabularx}
\usepackage{ragged2e}
\newcolumntype{P}[1]{>{\RaggedRight\hspace{0pt}}p{#1}}
\newcolumntype{Z}{>{\centering\let\newline\\\arraybackslash\hspace{0pt}}X}

\newcommand\Tstrut{\rule{0pt}{2.5ex}}         
\newcommand\Bstrut{\rule[-1.0ex]{0pt}{0.5pt}}   

\newtheorem*{setup}{Problem Setup}

\begin{document}

\title{
	 Reactive Motion Generation With Particle-Based Perception in Dynamic Environments
}

\author{	\medskip

Xiyuan Zhao, Huijun Li, Lifeng Zhu, Zhikai Wei, Xianyi Zhu, and Aiguo Song

\thanks{This work was supported by Fundamental and Interdisciplinary Disciplines Breakthrough Plan of the Ministry of Education in China JYB2025XDXM206. (e-mail: xiyuan.zhao@seu.edu.cn; lihuijun@seu.edu.cn; a.g.song@seu.edu.cn)}
\thanks{The authors are with the State Key Laboratory of Digital Medical Engineering, Jiangsu Key Lab of Robot Sensing and Control, School of Instrument Science and Engineering, Southeast University, Nanjing 210096, China.}
}



\maketitle

\begin{abstract}
	
Reactive motion generation in dynamic and unstructured scenarios is typically subject to essentially static perception and system dynamics. Reliably modeling dynamic obstacles and optimizing collision-free trajectories under perceptive and control uncertainty are challenging. This article focuses on revealing tight connection between reactive planning and dynamic mapping for manipulators from a model-based perspective. To enable efficient particle-based perception with expressively dynamic property, we present a tensorized particle weight update scheme that explicitly maintains obstacle velocities and covariance meanwhile. Building upon this dynamic representation, we propose an obstacle-aware MPPI-based planning formulation that jointly propagates robot-obstacle dynamics, allowing future system motion to be predicted and evaluated under uncertainty. The model predictive method is shown to significantly improve safety and reactivity with dynamic surroundings. By applying our complete framework in simulated and noisy real-world environments, we demonstrate that explicit modeling of robot-obstacle dynamics consistently enhances performance over state-of-the-art MPPI-based perception-planning baselines avoiding multiple static and dynamic obstacles.

\end{abstract}


\IEEEpeerreviewmaketitle

\section{Introduction}
\label{Sec:Introduction}

\IEEEPARstart{R}{eactive} and sensor-based planning paradigms effectively enhance the capability to autonomous exploration and real-world physical human-robot interactions (pHRI), particularly for unstructured and dynamic environments \cite{choset2005principles, billard2019trends, azkarate2022design, ren2023robot, koptev2024reactive}. For real-world manipulation, reactive planning greatly benefits from the realtime perception to potentially dynamic obstacles. Recent researches have studied both autonomous planning and mapping for unmanned aerial vehicles (UAVs) or autonomous ground vehicles (AGVs). However, only a little attention has been paid to reactive perception and control for manipulators.

Most existing planning algorithms mainly focus on solving high-dimensional problems posed by manipulators with high degrees of freedom (DOF) \cite{khansari2012dynamical, zucker2013chomp, bhardwaj2022storm, koptev2024reactive}. These methods typically assumed that the manipulator operates in perfectly perceptive and known environments. 
However, the increased applications have been paid to unstructured settings with multiple static and dynamic obstacles. In such scenarios, it is required for robots to realize realtime perception to unstructured environments and avoid obstacles agilely. An alternative line of the research aims to endow manipulators with explicit environmental awareness by sensing and reconstructing the surroundings with voxels or point cloud using on-board sensors \cite{ren2023robot, liu2024manipulability, border2024surface, breyer2021volumetric}. While these approaches enable active perception, the associated perception-planning pipelines tend to be offline, limiting applications to static and unstructured environments. In this work, we bridge the gap between realtime mapping and motion generation by introducing a tightly coupled, sensor-based control framework for manipulators in dynamic environments.  

Tight coupling between real-time mapping and reactive motion planning provides a powerful paradigm for robotic tasks. Several prior works \cite{kappler2018real, power2024learning, zhu2023real} have evaluated the integration of the raycast-based perception with reactive motion generation, preliminarily demonstrating its potential benefits. 
However, the dynamic characteristics of unstructured environments has not been explicitly considered and sufficiently explored.
To focus on coupling principles between real-time perception and reactive control, we consider a generalized sensing setup with both on-board and fixed RGB-D sensor in method analysis.
Under this formulation, we study the collision-free motion generation problem for robot manipulators in dynamic and unstructured environments. Recent advances in perception have introduced particle-based maps \cite{chen2024continuous, min2021kernel} in 3-D dynamic environments for UAVs and AGVs. Due to the ability to model dynamic and arbitrary-shaped obstacles, such map is naturally compatible with model predictive control, thereby providing an insightful foundation for the perception–planning framework considered in this work. Despite the promising direction, there are three main challenges: (1) 
Articulated manipulators are subject to high-dimensional kinematic constraints, arm geometries in the global frame and real-time requirements, which necessitate a global map tailored to the manipulation task.
(2) Restricted by perception used before, most reactive planning for manipulators only consider collision avoidance in the static or 
structured environments. In dynamic scenarios, the dynamics propagation for the manipulator and obstacles need to be further handled. (3) While many works have addressed real-time perception or planning individually, the unified integration of depth sensor feedback into motion generations, especially for manipulators in dynamic and uncertain environments, remains limited.

In this article, we propose a novel sensor-based perception and planning framework for manipulators, called 
\textbf{S}imultaneous \textbf{M}apping \textbf{A}nd \textbf{R}eactive Planning Using \textbf{T}ensor Optimization (SMART), which introduces particle-based perception (maps) to generate reactive collision-free motion in unstructured and dynamic environments. With the aim of maximizing reactive performance to dynamic obstacles, we develop a global continuous occupancy map for a cuboid manipulation space in the batch-operated fashion. The Model Predictive Path Integral (MPPI)-based control is generalized to react to dynamic obstacles perceived by our maps. Different from either end-to-end methods or those for point-like robots, we focus on high-DOF robot manipulators with the theoretically interpretable nature. Also, we aim to reveal the link between mapping and planning in uncertain and dynamic environments. The contributions of our work are summarized as follows:

\begin{itemize}
	
	\item[1)] We propose a global particle-based continuous occupancy map namely global dual-structure particle-based (G-DSP) map for arm manipulations with explicit uncertainty and velocity modeling in dynamic environments.
	\item[2)] We extend MPPI formulation to handle dynamic collision avoidance via propagating robot-obstacle dynamics and considering uncertainty of obstacles in manipulations.
	\item[3)] We propose a tensorized reactive and sensor-based planning paradigm combing G-DSP and MPPI-based control for manipulators in dynamic collision avoidance tasks.
	\item[4)] We develop a realtime implementation for SMART using graphics processing units (GPU), and empirically evaluate our framework in extensive experiments using UR5 robot, avoiding multiple dynamic and static obstacles. The open source code of G-DSP map using CUDA will be released.
	
\end{itemize}

The rest of this paper is organized as follows: Section \ref{Sec:RelatedWork} and \ref{Sec:Background} cover related works and background knowledge. In Section \ref{Sec:Approach1}, we present our proposed G-DSP for manipulation in detail. Section \ref{Sec:Approach2} describes the reactive planning method considering dynamic obstacles. Following this, our framework is evaluated through extensive simulation and real robot experiments. Finally, Section \ref{Sec:Conclusion} concludes the article and discusses future research directions.

\section{Related Work}
\label{Sec:RelatedWork}


\subsection{Perception and Mapping for Manipulation}

For articulated rigid manipulators, perception and mapping is important in sensor-based planning \cite{bohg2017interactive}. A typical realization is to approximate targets of interest via simple primitives, such as spheres, point cloud, voxels, signed distance functions (SDF), etc. For fully known scenarios, many works \cite{zucker2013chomp, kappler2018real, bhardwaj2022storm, hoerger2023multilevel, zhu2023real, huber2024avoidance, michaux2024safe, koptev2024reactive} observe surroundings with a global camera or multiple Optitrack trackers. In \cite{zucker2013chomp, huber2024avoidance, michaux2024safe}, the primitive SDFs are used to compute cost or modulate system dynamics. Using OctoMap \cite{hornung2013octomap}, \cite{kappler2018real} perceives obstacles with voxels in the limited workspace. \cite{bhardwaj2022storm} represents objects in static scenes as filtered point cloud at the start of run. \cite{palleschi2021fast} models human body through a real-time skeleton detection network. However, these works consider structured or static scenarios, which is not suitable to imperfect sensing in dynamic settings.

To tackle this problem, there are a few solutions using in-hand sensors. In static and unknown environments, authors of \cite{ren2023robot, liu2024manipulability, saund2023blindfolded} present effective offline perception and planning algorithms using visual or torque sensors. \cite{ren2023robot} maps cluttered cabinets through point cloud registration, realizing a collision-free planning with FCL \cite{pan2012fcl}. Such scheme is time-consuming due to collision querying for millions of raw points. To accelerate planning, voxel occupancy is widely used for manipulators \cite{liu2024manipulability, saund2023blindfolded}. Inspired by Nerf \cite{mildenhall2021nerf}, recent neural radiance field-based maps \cite{ortiz2022iSDF, chen2024catnips, liu2024physics} have emerged for high-fidelity dense rendering of precise manipulation with relatively low rate.

These studies focus on perception and planning in unstructured but static environments. A recent work \cite{abdelrahman2024neuromorphic} provides an event-based perception for manipulation avoiding moving obstacles, which takes the advantage of low power consumption but lacks complex stereo sensing. For most mapping methods, it is difficult to acquire dynamical characteristics of obstacles or suffer from distinguishing between static and dynamic parts. Fortunately, particle-based maps have paved a way to model obstacles with velocity estimation for AGVs and UAVs \cite{danescu2011modeling, chen2024continuous}, which is also promising for the high-dimensional reactive planning for manipulators. To the best of our knowledge, the study of an efficient integration between motion generation and particle-based maps via RGB-D sensors is novel.

\subsection{Particle-Based Dynamic Occupancy Maps}

The particle-based maps are characterized by modeling the surroundings with lots of weighted particles for point objects \cite{danescu2011modeling}. For occupancy estimation of 2-D grid cells, \cite{nuss2018random} proposes a probability hypothesis density/multi-instance Bernoulli filter, and realizes a realtime algorithm with parallel GPU. Then, \cite{min2021kernel} employs Bayesian generalized kernel inference algorithm and a counting sensor model to particle-based maps, extending dynamic occupancy maps to 3-D space.  Recently, DSP map \cite{chen2024continuous} models all point objects as particles, utilizing a dual structure representation to accelerate update and solving inconsistency caused by sparse measurements. Also, \cite{han2023ds} and \cite{chen2024continuous} improve the dynamic performance using Dempster-Shafer theory.

While a few effective particle-based implementations have been proposed, these maps can not be used directly to reactive planning of manipulators. Assuming a mobile or aerial robot to be a spherical mass point, the DSP map is egocentric in the center of cuboid. Due to the essence of articulated body, it is important to model and predict surroundings using a preferable particle-based map. To this end, we aim to introduce the serial robot kinematics and geometries, creating a global map for the limited manipulation space in the world frame. Moreover, we propose a batch operation for particles to accelerate mapping in order to meet realtime requirements. Rather than resolving inconsistency of sparse point cloud, we use particles to model obstacles in FOV to estimate dynamics of primitives naturally.

\subsection{Reactive Motion Generation With Perception}

Reactive planning has been widely studied through different research methods in recent decades. To guarantee realtime convergence for dynamic surroundings, the field-based planning methods are developed analytically. The commonly used Artificial Potential Fields (APF) \cite{khatib1986real, tulbure2020closing} and Vector Fields (VF) \cite{goncalves2010vector, becker2024motion} generate the control field for dynamic obstacles. To solve local minima and concave-shaped obstacle problem, \cite{huber2024avoidance} expands Dynamical System Modulation (DSM) \cite{khansari2012dynamical} by rotating dynamics toward the tangent space with structured perception. However, such methods require exact geometry representation or simple shapes, making it challenging to extend to the joint space with high dimension. Another popular approach is tree-based method. Many classical reactive planning methods \cite{thrun2005probabilistic, otte2016rrtx, rajendran2021human} were proposed to deal with dynamic environments. For uncertain robot state, Monte Carlo Tree Search (MCTS)-based method partially observable Monte-Carlo planning (POMCP) \cite{silver2010monte} and its variants \cite{cai2021hyp, hoerger2024adaptive, ragan2024online} have been proven to be valid for planning in belief space. Moreover, Bai et al. \cite{bai2014integrated} propose an integrated perception and planning framework in continuous belief space for AGVs. Through robot state, action sampling and observation reception, the optimal control will be executed by rollouting the promising part of belief tree. Optimization-based methods via cost functional minimization \cite{zucker2013chomp, usenko2017real} and motion primitives \cite{carvalho2025motion} in static scenarios are also widely used.

In numerous high-dimensional planning tasks of rigid body, robot states from encoders are accurate and state uncertainty is negligible. For such cases, model predictive control (MPC) is an efficient framework. The simplification from belief space to state space accelerates simulation and focuses on sampling the optimal control with regard to cost function under control uncertainty. Despite the similar objective, different optimization processes led to the development of two solutions: robust MPC (RMPC) and stochastic MPC (SMPC). For the former, RMPC optimize control costs with hard constraints, leading to overly conservative action for complex environments \cite{bemporad2007robust, blackmore2011chance}. \cite{lindqvist2020nonlinear, rastgar2024priest} consider obstacles as deterministic constraints and explicitly push trajectories the safe region. For manipulators, \cite{zhu2023real} proposed a cascaded RMPC and utilized a global camera to estimate velocity of an obstacle with super-twisting observer. In contrast, SMPC considers uncertainty with probability theory. Some works \cite{du2011probabilistic, zhu2019chance, de2024topology} adopted chance constraints to improve performance of collision checking in dynamic environments. 

As another popular SMPC, model predictive path integral (MPPI) theory solves the stochastic optimal control problems with Monte Carlo simulation and weighted sum for sampled control trajectories with the implicit constraints driven by cost functions. The idea of path integral is initially applied to the optimal control \cite{PhysRevLett.95.200201} and the reinforcement learning \cite{theodorou2010generalized, kalakrishnan2011stomp} under the assumption of control affine. Then, Williams et al. \cite{williams2017information, williams2018information} generalized MPPI to complex nonlinear dynamics from the information-theoretic perspective. Subsequent MPPI-based planning methods are mainly developed to improve its action sampling \cite{mohamed2022autonomous, vlahov2024low} or incorporate state uncertainty \cite{mohamed2025towards} for high-speed AGVs or UAVs. Among these works, Bhardwaj et al. \cite{bhardwaj2022storm} extended MPPI to the high dimensional joint space with pre-computed static world model through approximated dynamics modeling and parallel GPU speedup. Based on MPPI \cite{williams2018information}, \cite{koptev2024reactive} proposed an efficient planning framework combined MPPI sampling and DSM to overcome local minima in joint space, avoiding a set of dynamic spherical obstacles marked beforehand through increasing control rate. However, none of such research studied MPPI for manipulators in unstructured and dynamic environments. Also, dynamical propagation for complex-shaped obstacles has not been yet considered. Thus, a real-world implementation of MPPI-based manipulation in dynamic environments remains a challenge.

\section{Preliminaries}
\label{Sec:Background}

In this section, we briefly introduce core concept Sequential Monte-Carlo probability hypothesis density (SMC-PHD) filter used in mapping. For the detailed description, please refer to \cite{danescu2011modeling, ristic2013particle, chen2024continuous}. Next, we describe the MPPI-based parallelized reactive control method STORM \cite{bhardwaj2022storm}, which is the basis of our planning module in high-dimensional joint spaces.
Finally, we provide the system overview of proposed framework SMART. As this work involves both mapping and planning, we give the detailed definition for each variable. To maintain conventions of naming mapping and control variables, the bold italics ($\bm{x}$) and boldface ($\mathbf{x}$) are used for distinctions. Lower-case symbols ($x$), bold lower-case symbols ($\bm{x}$ or $\mathbf{x}$) and upper-case ones ($X$) denote scalars, vectors and matrices (or tensors), respectively. Several upper-case symbols (like $L$, $H$) are used for constants. 

\begin{figure*}
	\centering
	\setlength{\abovecaptionskip}{0.15cm}
	\includegraphics[width=0.96\linewidth]{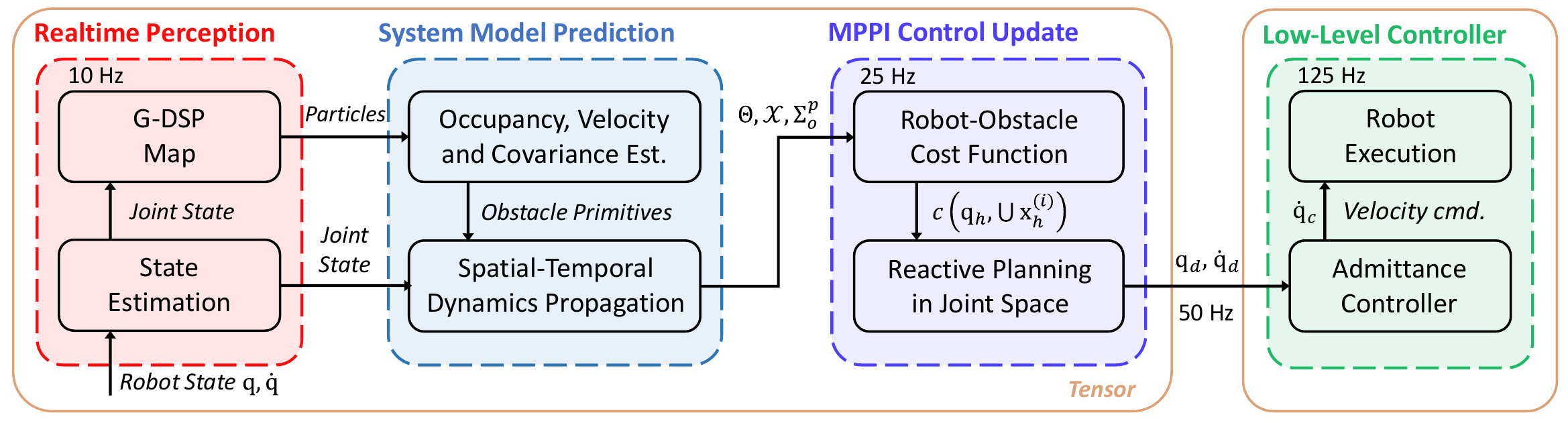}
	\caption{The pipeline of our proposed sensor-based planning framework SMART, including the realtime perception (dynamic occupancy mapping), dynamics propagation and reactive motion generation components.}
\label{fig:Pipeline}
\end{figure*}

\subsection{SMC-PHD Filter}

Considering a random finite set (RFS) defined in state space of a map $\mathbb{M}$
\begin{equation}
	\mathrm{X} = \left\{ \bm{x}^{(1)}, \bm{x}^{(2)}, ..., \bm{x}^{(N)} \right\},
\end{equation}
where a RFS is a finite set-valued variable representing states of tracked objects, $\bm{x}^{(i)} = (p_x, p_y, p_z, v_x, v_y, v_z) \in \mathbb{R}^{6}$ is the state vector in the RFS, including 3D position and 3D velocity components of $i$-th point object. Subscripts $x$, $y$, $z$ indicate three Cartesian axis. We will also use $\bm{p}^{(i)} = (p_x, p_y, p_z) \in \mathbb{R}^3$ and $\bm{v}^{(i)} = (v_x, v_y, v_z) \in \mathbb{R}^3$ to denote 3-D position and velocity. Note that $|\mathrm{X}| = N \in \mathbb{N}$ is random variable estimating object number in $\mathrm{X}$ which is called cardinality. The probability hypothesis density of $\mathrm{X}$ is defined as its first moment, i.e.
\begin{equation}
	D_{\mathrm{X}}(\bm{x}) = \mathbb{E}_{\mathrm{X}}\left[ \sum_{i = 1}^{N} \delta (\bm{x} - \bm{x}^{(i)}) \right],
\end{equation}
where $\delta$ is the Dirac function.

The core of the particle-based map is probability hypothesis density (PHD) filter. In order to efficiently realize PHD filter, Sequential Monte-Carlo PHD (SMC-PHD) filter is proposed to approximate PHD with weighted particles. With $L_{t-1}$ particles $\tilde{\bm{x}}^{(i)}$, $(i = 1, ..., L_{t-1})$ at $t-1$, the posterior PHD is given by
\begin{equation}
	D_{\mathrm{X}_{t-1}}(\bm{x}) \approx \sum_{i = 1}^{L_{t-1}} \omega_{t-1}^{(i)} \delta(\bm{x} - \tilde{\bm{x}}^{(i)}),
\end{equation}
where $\omega_{t-1}^{(i)}$ denote weight of particle $i$. The filtering process is divided into two steps. Firstly, given the transition function $\pi_{t|t-1}$ for each object, the prediction equation is derived as
\begin{align}
	D_{\mathrm{X}_{t|t-1}}(\bm{x}) &= D_{\mathrm{S}_{t|t-1}}(\bm{x}) + D_{\mathrm{B}_{t|t-1}}(\bm{x}) \notag \\
	&= P_{s} \sum_{i = 1}^{L_{t-1}} \omega_{t-1}^{(i)} \pi_{t|t-1}(\bm{x}|\tilde{\bm{x}}_{t-1}^{(i)}) + \gamma_{t|t-1}(\bm{x}),
	\label{prediction1}
\end{align}
where the prediction consists of two parts: the PHD of survived objects propagated from the last time step $D_{\mathrm{S}_{t|t-1}}(\bm{x})$ and the PHD of RFS for newborn particles $D_{\mathrm{B}_{t|t-1}}(\bm{x})$. Here, $D_{\mathrm{B}_{t|t-1}}$ is modeled by Poisson point process with intensity $\gamma_{t|t-1}$. $P_s$ is the probability of survival from $t-1$. Let $\omega_{s,t|t-1}^{(i)} \triangleq P_{s} \omega_{t-1}^{(i)}$. After sampling particles from Eq. \eqref{prediction1}, we have
\begin{align}
	D_{\mathrm{X}_{t|t-1}} &= \sum_{i = 1}^{L_{t-1}} \omega_{s, t|t-1}^{(i)} \delta (\bm{x} - \tilde{\bm{x}}_{s, t|t-1}^{(i)}) + \sum_{j = 1}^{L_{b, t}} \omega_{b, t}^{(j)} \delta (\bm{x} - \tilde{\bm{x}}_{b, t}^{(j)}) \notag \\
	&= \sum_{i = 1}^{L_{t}} \omega_{t|t-1}^{(i)} \delta (\bm{x} - \tilde{\bm{x}}_{t}^{(i)}),
	\label{prediction}
\end{align}
where the number of predicted particles is $L_{t} = L_{t-1} + L_{b, t}$. 

Real-world measurements are inevitably noisy. Thus, SMC-PHD filter assumes that objects will be detected with a probability $P_d$ and update via likelihood $G(\bm{z}_{t}, \bm{x})$. The unobserved objects are preserved with $1 - P_d$. Upon receiving observation $\mathrm{Z}_{t}$, the update is computed as
\begin{equation}
	D_{\mathrm{X}_{t}}(\bm{x}) = \left[ 1 - P_d + P_d \sum_{\bm{z}_{t} \in \mathrm{Z}_{t}} G(\bm{z}_{t}, \bm{x}) \right] D_{\mathrm{X}_{t|t-1}}(\bm{x}),
	\label{update1}
\end{equation}
where the likelihood is given by
\begin{equation}
	G(\bm{z}_{t}, \bm{x}) = \frac{\mathrm{g}(\bm{z}_{t} | \bm{x})}{\kappa(\bm{z}_t) + P_d \int \mathrm{g}(\bm{z}_{t} | \bm{x}) D_{\mathrm{X}_{t|t-1}} \mathrm{d}\bm{x}},
	\label{likelihood}
\end{equation}
where $\kappa(\bm{z}_{t})$ is the PHD of false detection, and $\mathrm{g}(\bm{z}_{t}, \bm{x})$ is the likelihood for each object. Combining \eqref{prediction} and \eqref{update1}, we have 
\begin{align}
	D_{\mathrm{X}_{t}} &= \sum_{i = 1}^{L_{t}} \left[ 1 - P_d + P_d \sum_{\bm{z}_{t} \in \mathrm{Z}_{t}} G(\bm{z}_{t}, \tilde{\bm{x}}_{t}^{(i)}) \right] \omega_{t|t-1}^{(i)} \delta(\bm{x} - \tilde{\bm{x}}_{t}^{(i)}) \notag \\
	&= \sum_{i = 1}^{L_{t}} \omega_{t}^{(i)} \delta(\bm{x} - \tilde{\bm{x}}_{t}^{(i)}),
	\label{update}
\end{align}
where $\omega_{t}^{(i)} \triangleq \left[ 1 - P_d + P_d \sum_{\bm{z}_{t} \in \mathrm{Z}_{t}} G(\bm{z}_{t}, \tilde{\bm{x}}_{t}^{(i)}) \right] \omega_{t|t-1}^{(i)}$. In brief, SMC-PHD is predicted using \eqref{prediction} and updated according to Eq. \eqref{update} iteratively.

\subsection{MPPI and STORM Formulation}
\label{storm}

Considering a discrete-time stochastic dynamical system of a $d$-DOF manipulator, which evolves in the state space $\mathcal{S}$ under given control space $\mathcal{U}$ and with dynamics $f$ given by
\begin{equation}
	\mathbf{q}_{t+1} = f(\mathbf{q}_{t}, \mathbf{w}),
\end{equation}
where the current state $\mathbf{q}_{t} \in \mathcal{S} \subset \mathbb{R}^{d}$ denotes joint position, and the control input is a random variable satisfying Gaussian distribution $\mathbf{w} = \mathbf{u} + \delta \mathbf{u} \sim \mathcal{N}(\mathbf{u}, \tilde{\Sigma}_{c}) \in \mathcal{U}$, with an initial control $\mathbf{u}$ and fixed covariance $\tilde{\Sigma}_{c} \in \mathbb{R}^{d \times d}$. For manipulators, we define the control term as the joint accelerations $\mathbf{w} \triangleq \ddot{\mathbf{q}}_{d} \in \mathbb{R}^d$. The canonical MPPI \cite{williams2018information} solves the optimal control problem via three key steps: 1) At each time step, $K \in \mathbb{N}^{+}$ control trajectories $\{V_k\}_{k = 1}^{K} = \{(\mathbf{w}_{0}, \mathbf{w}_{1}, ..., \mathbf{w}_{H-1})\}_{k = 1}^{K} \in \mathbb{R}^{K \times H \times d}$ along horizon $H$ are sampled from nominal control sequence $U = (\mathbf{u}_{0}, \mathbf{u}_{1}, ..., \mathbf{u}_{H-1}) \in \mathbb{R}^{H \times d}$ and are propagated through approximated dynamical model $\tilde{f}$ to derive $K$ state trajectories $\{\tau_k\}_{k = 1}^{K} = \{(\mathbf{q}_{1, k}, \mathbf{q}_{2, k}, ..., \mathbf{q}_{H, k})\}_{k = 1}^{K}$. 2) After that, the cost of each trajectory $c(\tau_k)$ is calculated. 3) The optimal control sequence $U^{*} \in \mathbb{R}^{H \times d}$ is obtained by the weighted sum of $K$ sequences along horizon, sending the first control to robot and leaving $H - 1$ items to warm-start the next iteration. 

The improvement upon MPPI for high-dimensional manipulation is STORM \cite{bhardwaj2022storm}. Instead of using a constant control covariance, STORM utilizes a more flexible adaptive covariance update rule, which is defined as
\begin{align}
		\eta_{k} &\propto \exp \left( -\frac{1}{\beta} \left(c(\tau_{k}) - \underset{i = 1, ..., K}{\min} c(\tau_{i}) \right) \right), \label{storm_equ1} \\
		U &= \left(1 - \alpha_{u}\right) U + \alpha_{u} \sum_{k = 1}^{K} \eta_{k} V_{k}, \label{storm_equ2} \\
		\Sigma_{c} &= \left(1 - \alpha_{\Sigma_c}\right) \Sigma_{c} + \alpha_{\Sigma_c} \sum_{k = 1}^{K} \eta_{k} \left(V_{k} - U\right)^{\top} \left(V_{k} - U\right),
		\label{storm_equ3}
\end{align}
where $\eta_{k}$ is the normalized weight of $k$-th trajectory, $\beta \in \mathbb{R}^{+}$ is the inverse temperature coefficient. The terms $\alpha_{u}$ and $\alpha_{\Sigma_c}$ are step sizes to regulate mean and covariance. As mentioned in \cite{bhardwaj2022storm}, STORM utilizes a sample-based gradient approach to update control. Compared with MPPI, STORM warms-starts the control mean and covariance in the horizon simultaneously. While STORM accelerates the control loop to 50-100 Hz, it is still difficult to deal with desired dynamic and unstructured environments, which is our focus in this work.

\subsection{System Overview}

The overall pipeline of the SMART framework is illustrated in Fig. \ref{fig:Pipeline}. In this work, we aim to generate reactive collision-free motion between the initial and target configurations with dynamic obstacles. To better deploy to real-world robot applications, we consider environments where there exists multiple arbitrarily shaped dynamic or static obstacles. To avoid these obstacles in time, we develop a MPC-based reactive planning algorithm in joint space namely D-STORM, propagating the dynamics of both robot and obstacles in parallel (see Section \ref{Sec:Approach2}). The rollout of obstacles relies on our stereo particle-based perception, occupancy and velocity estimation of G-DSP map. The details of G-DSP map are presented in Section \ref{Sec:Approach1}. Due to the stochastic nature of G-DSP map for manipulators, the positions, velocities and covariance of obstacle primitives can be calculated naturally. Then, spatial-temporal system dynamics propagation and uncertainty-aware costs are accumulated along planning horizon. Another unique aspect of our approach is that the synergy between global mapping and planning is realtime using tensor optimizations on GPUs. The advantage of parallelized tensor operation facilitates the high-frequency perception and control with low latency. This ensures real-time perception and reaction to the surroundings.

Given limited FOV of depth camera used in manipulations, we can enforce the following assumption for the \textit{unstructured} environments in our subsequent study. Under the Assumption of manipulation in unstructured surroundings, there would be the theoretical guarantee that a robot will be not at the risk of collision beyond its FOV in theory.

\begin{setup}
	(Trivially Unstructured Environment) In this work, all static and moving obstacles are assumed to be impenetrable, and are fully observable by robots (depth sensors).
\end{setup}

\section{Global Particle-based Mapping For Manipulators}
\label{Sec:Approach1}

In dynamic environments, the robot is anticipated to observe positions, velocities, uncertainty of static and moving obstacles to react in time. Therefore, we propose a particle-based map based on the global frame, which incorporates the kinematics and geometries of the manipulator and obstacles in the limited and fixed workspace. Based on batch operations, a tensorized paradigm of the particle-based map is introduced.

\subsection{Global Dual-Structure Map}
\label{subsec1}

For a $d$-DOF manipulator $\mathcal{R}$, We set up the world coordinate system with the base as origin $O$. The pose of the end-effector $\mathbf{x}_{E} \in \mathbb{R}^{6}$ can be found using forward kinematics relationship
\begin{equation}
	\mathbf{x}_{E} = \mathrm{FK}\left( \mathbf{q} \right),
	\label{fk}
\end{equation}
where 
$\mathbf{q} \in \mathbb{R}^{d}$ is current robot joint position. The global pose of fixed camera $\mathbf{T}_{wc}$ can be calibrated. When the camera is mounted on the end-effector with a fixed pose $\mathbf{T}_{ec} \in \mathbf{SE}(3)$. From \eqref{fk}, we can also derive the pose by an end-effector-to-world transform $\mathbf{T}_{we}$ in the world frame, i.e. $\mathbf{T}_{wc} = \mathbf{T}_{we} \mathbf{T}_{ec}$. Thus, for each observation $\bm{z}_{t} \in \mathrm{Z}_{t}$ in point cloud we have
\begin{equation}
	\left[ \begin{array}{c}
		\tilde{\bm{z}}_{t} \\ 1
	\end{array} \right] = \mathbf{T}_{wc} \left[ \begin{array}{c}
	\bm{z}_{t} \\ 1
\end{array} \right],
\end{equation}
where $\tilde{\bm{z}}_{t} \in \mathbb{R}^{3}$ is the observed point in the world frame. Let $\partial \mathcal{R}(\mathbf{q})$ denote the surface of manipulator. The signed distance function can be defined as $\mathrm{dist}(\bm{p}) = \pm \inf_{\bm{p}' \in \partial \mathcal{R}(\mathbf{q})} \Vert \bm{p} - \bm{p}' \Vert_{2}$, which is the minimum distance between a point $\bm{p}$ and robot surface at configuration $\mathbf{q}$. In our map, observations inside or near the surface of robot will be ignored, $\mathrm{dist}(\tilde{\bm{z}}_{t}) \leq \epsilon_d$ with $\epsilon_d > 0$. The G-DSP map will be discussed in the world frame. 

The occupancy status is estimated using the number of point objects at arbitrary locations in the particle-based map. For an arbitrary voxel resolution $l$, we use voxel filter to handle raw point cloud. Using \eqref{update}, the expectation of cardinality for RFS in $k$-th voxel $\mathbb{V}_k$ can be derived as
\begin{equation}
	\mathbb{E}_{\mathrm{X}_{t}}[|\mathrm{X}_{t}^{\mathbb{V}_k}|] = \int D_{\mathrm{X}_{t}^{\mathbb{V}_k}} (\bm{x}) \mathrm{d} \bm{x} = \sum_{\tilde{\bm{x}}_{t}^{(i)} \in \mathbb{V}_{k}} \omega_{t}^{(i)},
\end{equation}
i.e. the sum of weights for all particles in the voxel. Then, the occupancy probability of $\mathbb{V}_k$ is estimated by
\begin{equation}
	P_{occ}(\mathbb{V}_k) = \mathbb{E}_{\mathrm{X}_{t}}[|\mathrm{X}_{t}^{\mathbb{V}_k}|].
	\label{occupancy_estimation}
\end{equation}
Given a proper threshold, the binary occupancy status (free or occupied) is obvious. Then, the center position of occupancy $\mathbf{p}(\mathbb{V}_k) \in \mathbb{R}^3$ can be easily derived. Moreover, the velocities of occupied voxels $\mathbf{v}\left( \mathbb{V}_k \right) \in \mathbb{R}^3$ can be further estimated as the weighted average of inner particle velocities
\begin{equation}
	\mathbf{v}\left( \mathbb{V}_k \right) = \sum_{\tilde{\bm{x}}_{t}^{(i)} \in \mathbb{V}_{k}} \tilde{\omega}_{t}^{(i)} \tilde{\bm{v}}_{t}^{(i)},
	\label{velocity_estimation}
\end{equation}
where $\tilde{\bm{v}}_{t}^{(i)}$ is the estimated 3-D velocity of the particle $\tilde{\bm{x}}_{t}^{(i)}$, $\tilde{\omega}_{t}^{(i)}$ denotes normalized weight of particle $\tilde{\bm{x}}_{t}^{(i)}$ in its voxel. Clearly, those particles with larger weights have the greater impact on the voxel velocity. The uncertainty of voxel position and velocity estimation can be explicitly modeled by Gaussian covariance, which is given by
\begin{align}
	\Sigma_o^\mathrm{p} (\mathbb{V}_k) &= \sum_{\tilde{\bm{x}}_{t}^{(i)} \in \mathbb{V}_{k}} \tilde{\omega}^{(i)}_t \left( \tilde{\bm{p}}^{(i)}_t - \mathbf{p}(\mathbb{V}_k) \right) \left( \tilde{\bm{p}}^{(i)}_t - \mathbf{p}(\mathbb{V}_k) \right)^\top, \notag \\
	\Sigma_o^\mathrm{v} (\mathbb{V}_k) &= \sum_{\tilde{\bm{x}}_{t}^{(i)} \in \mathbb{V}_{k}} \tilde{\omega}^{(i)}_t \left( \tilde{\bm{v}}^{(i)}_t - \mathbf{v} \left( \mathbb{V}_k \right) \right) \left( \tilde{\bm{v}}^{(i)}_t - \mathbf{v} \left( \mathbb{V}_k \right) \right)^\top,
	\label{covariance_estimation}
\end{align}
where $\tilde{\bm{p}}^{(i)}_t$ is the 3-D position of particle $\tilde{\bm{x}}_{t}^{(i)}$. For those static voxels, there exist $\mathbf{v} \left( \mathbb{V}_k \right) = 0$ and $\Sigma_o^\mathrm{v}\left( \mathbb{V}_k \right) = \mathbf{0}$. Now our task is to estimate PHD using SMC-PHD filter. Before performing filtering, the map is partitioned for computational efficiency.

\begin{figure*}
	\centering
	\setlength{\abovecaptionskip}{0.00cm}
	\includegraphics[width=1.0\linewidth]{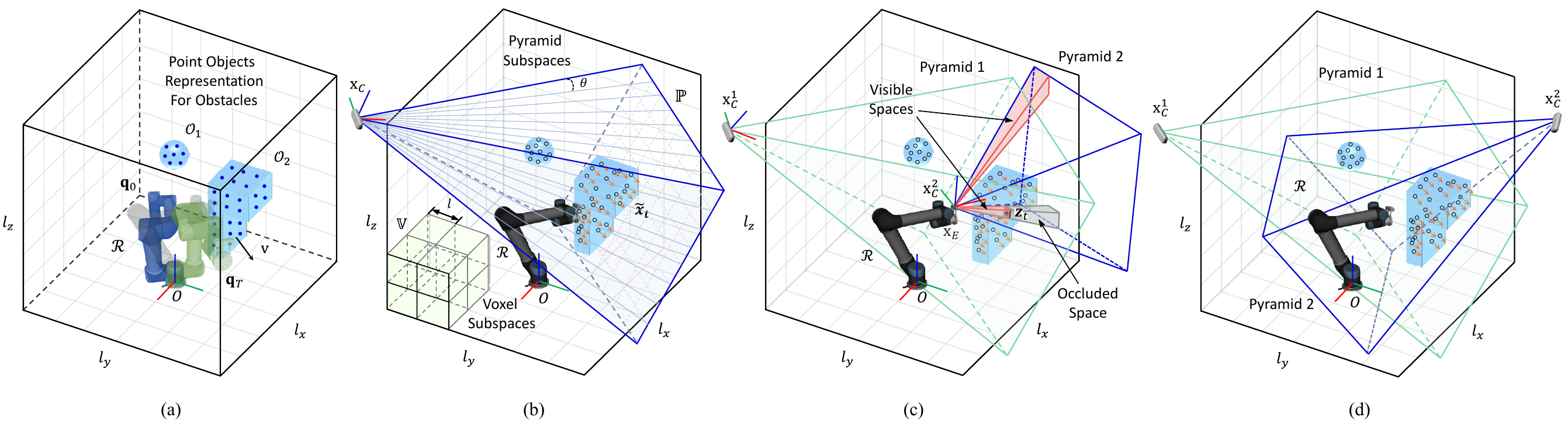}
	\caption{Visualization of G-DSP map. Since realtime planning for serial rigid robots involves the whole body, we consider the fixed base as the origin of world frame $O$. Thus, our map is global in the limited operation space. (a) Two obstacles of arbitrary shape are represented by point objects (blue solid circle). The manipulator starts from the blue initial configuration and must reach the green goal position while avoiding static and dynamic obstacles. (b) A fixed sensor setup. The obstacles are approximated by weighted particles (hollow circles) $\tilde{\bm{x}}_{t}$ with estimated positions and velocities. The mapping space are divided into two subspaces, i.e. voxel subspaces $\mathbb{V}$ and pyramid subspaces $\mathbb{P}$. (c) The dual-view setup with a fixed and an onboard camera for arm manipulation. The observations $\bm{z}_{t}$ from onboard camera are point clouds, which divides the pyramid space $\mathbb{P}$ into visible and occluded spaces. In narrow cases, a depth sensor fixed on $\mathbf{x}_C^1$ and an onboard camera mounted on robot $\mathbf{x}_C^2$ broadens perception for robots. (d) Another unstructured setup is to use two fixed cameras at $\mathbf{x}_C^1$, $\mathbf{x}_C^2$ to mitigate occlusion and realize whole-body collision avoidance. Our parallel global map works with common multi-view manipulation setups. }
	\label{fig:Mapping}
\end{figure*}

Like \cite{chen2024continuous}, we model point objects as a RFS in particle level and use two subspaces to realize filter spontaneously. Suppose mapping space $\mathbb{M}$ is cuboid with length $(l_x, l_y, l_z)$. For global mapping, we divide the whole map into $N_v = \frac{l_x \cdot l_y \cdot l_z}{l}$ voxel subspaces without overlap. Here, $l$ is called voxel resolution. We use $\mathbb{V}_{i}$ here to denote the $i$-th voxel subspace. Since voxel subspaces are fixed and $\mathbb{M} = \mathbb{V}$, $\mathrm{X}_{t} = \mathrm{X}_{t}^{\mathbb{V}_{1}} \cup \cdots \cup \mathrm{X}_{t}^{\mathbb{V}_{N_v}}$ is satisfied. Then we have
\begin{equation}
	D_{\mathrm{X}_{t}} (\bm{x}) = D_{\mathrm{X}_{t}^{\mathbb{V}_{1}}}(\bm{x}) + \cdots + D_{\mathrm{X}_{t}^{\mathbb{V}_{N_v}}}(\bm{x}) = \sum_{i = 1}^{N_v} D_{\mathrm{X}_{t}^{\mathbb{V}_{i}}}(\bm{x}).
\end{equation}
This is the fundamental that we use voxel subspace for particle resampling and occupancy estimation. Suppose the maximum number of particles allowed in G-DSP is $L_{\max}$, then there are at most $\frac{L_{\max}}{N_v}$ particles for each voxel subspace.

Meanwhile, for a camera with field of view (FOV) $\theta_h \times \theta_v$, the observation space $\mathbb{P}$ is divided into $N_p = \frac{\theta_h \times \theta_v}{\theta^2}$ pyramid subspaces $\mathbb{P}_{i}$ ($i = 1, ..., N_p$) with angle resolution $\theta$. Considering varying observation setups in real, two subspaces may not always overlap, see Fig. \ref{fig:Mapping} (b)-(d). Thus, only particles in the intersection space $\mathbb{P} \cap \mathbb{M}$ will be updated using point cloud, whereas those outside $\mathbb{P}$ but still in map $\mathbb{M}$ will be attenuated. Considering the maximum number of filtered points $M_{\max}$ in FOV pyramids, we approximate that the maximum observation distance in each pyramid subspace is equal. In Fig. \ref{fig:Mapping}, (a)--(c) illustrate the detailed representation of point objects and dual structure division of fixed and onboard cases. For the whole-body collision avoidance considered in this work, Fig. \ref{fig:Mapping} (d) visualizes unstructured settings in experiments. The algorithm of dual-view map (c)-(d) is shown in Appendix.

\subsection{Tensor Representation for Dual Subspaces}
\label{subsec2}

In SMC-PHD filter, predict and update operations both act on the particles in map. Inspired by the fact, we store states of all particles in an unique particle tensor while putting required partial states of particles into tensors of two subspaces. Thus, it is important to establish the relationship between them. For each particle, we use a state vector including 3-D position, 3-D velocity, weight, voxel index, pyramid index and state flag to express, i.e. $\tilde{\bm{x}} = (p_x, p_y, p_z, v_x, v_y, v_z, \omega, id_v, id_p, s)$, where $s$ is a flag indicating whether the particle is valid, $id_v$ and $id_p$ denote indices of voxel and pyramid subspaces respectively. The particle tensor is represented by $\tilde{X} \in \mathbb{R}^{L_{\max} \times 10}$, with the count of valid particle recorded to index the current states of particles from the particle pool $\tilde{X}$. To compute the maximum distance for observations in the map, $k$-th element in pyramid tensor is expressed by $\mathbb{P}_{k} = (L_{\mathbb{P}_k}, id_{f, \mathbb{P}_{k}}, id_{e, \mathbb{P}_{k}})$, where $L_{\mathbb{P}_k}$ is actual number of particles in $\mathbb{P}_k$, $id_{f, \mathbb{P}_{k}}$ and $id_{e, \mathbb{P}_{k}}$ are the first and last indices of particles in the pyramid sorted by pyramid indices respectively. Then the pyramid tensor is expressed by $\mathbb{P}_t \in \mathbb{R}^{N_p \times 3}$. To facilitate resampling and voxel estimation, the voxel tensor is modeled as $\mathbb{V}_t \in \mathbb{R}^{N_v \times L_{r} \times 10}$, storing all particle states into the corresponding voxel subspaces. Here, $L_{r}$ is the maximum number of survived and newborn particles in each voxel, and $L_r = 5 \frac{L_{\max}}{N_v}$ is set in our map. 

After sorting all particles according to pyramid index $id_p$, the next step is to assign the first index and the last index of particles in the same pyramid to each element in pyramid tensor. To efficiently find all pairs of indices, we circularly shift the pyramid index vector of all valid particles by one position and compare it with the original index vector, resulting in the first indices of each pyramid subspace with particles in particle tensor, denoted as $id_{f, \mathbb{P}}$. Then, a backward roll operation is applied to the original pyramid indices vector to search the end indices tensor $id_{e, \mathbb{P}}$. The pseudo-code of assignment to pyramid subspaces is given in Alg. \ref{alg1}.

To select observable particles and update their weights, we also utilize a observation tensor $\mathbb{P}_{z}$ to compute the maximum observation distance. First, we assign each pyramid subspace the distances from its inside points $\bm{z}_t$ to the sensor, resulting to $\mathbb{P}_{z} \in \mathbb{R}^{N_p \times \frac{M_{\max}}{N_p}}$. If there is no point in a pyramid subspace, all $\frac{M_{\max}}{N_p}$ distances corresponding to the subspace are default value, i.e. 0. If the current actual number of points in a pyramid subspace is less than the maximum number $\frac{M_{\max}}{N_p}$, remaining elements of the pyramid subspace in the observation tensor are filled with zeros. After that, we maximize the tensor along its second dimension to obtain the maximum observation distance i.e. $d_{\max} \in \mathbb{R}^{N_p}$, where $d_{\max}[i] = \underset{j = 1, ..., M_{\max}/N_p}{\max} \mathbb{P}_z[i, j] $.

\begin{algorithm}[t]
	\label{alg1}
	\caption{Assign Particles to Pyramid Subspaces}
	\SetAlgoLined
	\DontPrintSemicolon
	\KwIn{Particle Tensor $\tilde{X}_{t}$}
	\KwOut{Pyramid Tensor $\mathbb{P}_{t}$}
	$\tilde{X}_{t} \leftarrow \mathbf{sort} ( \tilde{X}_{t}.id_{p} )$ \\
	$id_{f, \mathbb{P}} \leftarrow \mathbf{compare} (\tilde{X}_{t}.id_{p}, \mathbf{roll} (\tilde{X}_{t}.id_{p}, 1) )$ \\
	$id_{e, \mathbb{P}} \leftarrow \mathbf{compare} (\tilde{X}_{t}.id_{p}, \mathbf{roll} (\tilde{X}_{t}.id_{p}, -1) )$ \\
	\tcp*[l]{\footnotesize{Pyramid indices tensor with particles}}
	$id_{\mathbb{P}} \leftarrow \tilde{X}_t [\tilde{X}_{t}.id_{p} \neq \mathbf{roll} (\tilde{X}_{t}.id_{p}, 1)].id_{p}$ \\
	$\mathbb{P}_{t}[id_{\mathbb{P}}, 1] \leftarrow id_{e, \mathbb{P}} - id_{f, \mathbb{P}} + 1$ \\
	$\mathbb{P}_{t}[id_{\mathbb{P}}, 2] \leftarrow id_{f, \mathbb{P}}$ \\
	$\mathbb{P}_{t}[id_{\mathbb{P}}, 3] \leftarrow id_{e, \mathbb{P}}$
\end{algorithm}

Similarly, particle assignment to voxel tensor is based on the sorted voxel indices of particles. 
Note that dual subspace tensors will not be valid at the same time for sorting. Nonetheless, our dual-structure assignment method is still effective. Firstly, two subspaces are not accessed simultaneously in the G-DSP map. When updating particle weights, only pyramid tensor is used. And only the voxel tensor will be inquired in resampling and occupancy estimation. Furthermore, the particle pool will highly decrease the time complexity in prediction and update step. In the following, we describe predict and update in detail.

\subsection{Mapping Using Batch Operation}
\label{subsec3}

After defining dual subspaces and their tensor representation in the workspace, G-DSP map is realized according to SMC-PHD filter. To speed up mapping, we utilize batch operations to predict, update and resample particles in parallel. First, the prediction step models $L_{s, t} = L_{t-1}$ valid particles in voxel subspaces as constant velocity model, which is given by
\begin{equation}
	\tilde{X}_{s, t}^{\mathrm{pv}} = A \tilde{X}_{s, t - 1}^{\mathrm{pv}} + \bm{\zeta},
	\label{predict}
\end{equation}
where $\tilde{X}_s^{\mathrm{pv}} = [\tilde{\bm{x}}^{(1)}_s[: 6], ..., \tilde{\bm{x}}^{(L_{s, t})}_s[: 6]]^\top \in \mathbb{R}^{L_{s, t} \times 6}$ is the position and velocity tensor of all survived particles from the last time step $t-1$. The transition tensor $A \in \mathbb{R}^{L_{s, t} \times 6 \times 6}$ is expressed by $\left[ \begin{array}{cc}
	I_{3 \times 3} & \Delta t I_{3 \times 3} \\ \mathbf{0}_{3 \times 3} & I_{3 \times 3}
\end{array} \right]$, where the first dimension is extended to $L_{s, t}$ for alignment. Here, tensor $\bm{\zeta} \in \mathbb{R}^{L_{s, t} \times 6}$ is the random Gaussian noise for prediction, with each sampled from $\mathcal{N} (\mathbf{0}_6, \Sigma_{\mathrm{pred}})$. Then, subscripts of voxel subspaces and pyramid subspaces for these particles are recalculated in batch.

As described in Section \ref{subsec1}, we only update particles in the intersection space of FOV and the mapping space $\mathbb{P}~\cap~\mathbb{M}$. Suppose there are $L_{f, t}$ valid and observable particles in $\mathbb{P}~\cap~\mathbb{M}$. For $M_{z, t}$ filtered observation points $\tilde{Z}_{t} \in \mathbb{R}^{M_{z, t} \times 3}$, we model likelihood as Gaussian distribution, which is represented as
\begin{equation}
	\mathrm{g}(\tilde{\bm{z}}_{t}|\tilde{\bm{x}}_{t}) = \mathcal{N}(\tilde{\bm{p}}_{t}, \Sigma_{z}),
\end{equation}
where $\tilde{\bm{p}}_{t} = \left[I_{3 \times 3} ~ \mathbf{0}_{3 \times 7}\right] \tilde{\bm{x}}_{t} = \tilde{\bm{x}}_t[: 3]$ is the position of particle states, and $\Sigma_{z} \in \mathbb{R}^{3 \times 3}$ is observation covariance. To simplify computations, it is assumed that the noise of observation in each Cartesian direction is independent and equal to $\sigma_z \in \mathbb{R}$. After receiving observation tensor $\tilde{Z}_{t}$, we have
\begin{equation}
	\mathrm{g}(\tilde{Z}_t|\tilde{X}_{f, t}) = \frac{1}{(2\pi)^{\frac{3}{2}}\sigma_z^3} \exp{\left( -\frac{1}{2 \sigma_z^2} \Vert \tilde{Z}_{t} - \tilde{X}_{f, t}^{\mathrm{p}} \Vert_{2}^{2} \right)},
	\label{g_batch}
\end{equation}
where $\tilde{X}_{f, t}^{\mathrm{p}} = [\tilde{\bm{p}}^{(1)}_t, ..., \tilde{\bm{p}}^{(L_{f, t})}_t]^\top \in \mathbb{R}^{L_{f, t} \times 3}$ denotes the current position tensor of observable particles. In order to keep the dimension alignment for tensor operations, we argument $\tilde{X}_{f, t}^{\mathrm{p}}$ on the first dimension to $\mathbb{R}^{M_{z, t} \times L_{f, t} \times 3}$ and replicate $\tilde{Z}_t$ along the second dimension to $\mathbb{R}^{M_{z, t} \times L_{f, t} \times 3}$. Norm operator $\Vert \cdot \Vert_2$ computes the distance between each observation and particle. Thus, we have likelihood tensor as $\mathrm{g}(\tilde{Z}_t|\tilde{X}_{f, t}) \in \mathbb{R}^{M_{z, t} \times L_{f, t}}$. For every observation $\tilde{\bm{z}}_t \in \tilde{Z}_t$, we accumulate the likelihood effects of all particles in the map rather than those in adjacent pyramid spaces in standard DSP map. 

\begin{algorithm}[t]
	\label{alg2}
	\caption{Update}
	\SetAlgoLined
	\DontPrintSemicolon
	\KwIn{Particle Tensor $\tilde{X}_{t} (\tilde{X}_{s, t})$, Point Tensor $\tilde{Z}_t$, Camera Pose $\mathbf{x}_C$, Expected Thickness $\epsilon_H$} 
	\KwOut{Updated Particle Tensor $\tilde{X}_t$}
	$\mathbb{P}_t \leftarrow \mathbf{AssignParticlesToPyramids}(\tilde{X}_{t})$ ~  $\triangleright$ Alg. \ref{alg1} \\
	$d_{z} \leftarrow \Vert \tilde{Z}_{t} - \mathbf{x}_C[: 3] \Vert_{2}$ \\
	$\mathbb{P}_z \leftarrow \mathbf{AssignPointsToPyramids}(\tilde{Z}_{t}, d_z)$ \\
	\tcp*[l]{\footnotesize{Maximum observation distance tensor}}
	$d_{\max} \leftarrow \underset{j = 1, ..., M_{\max}/N_p}{\max} \mathbb{P}_z[:, j] \in \mathbb{R}^{N_{p}}$ \\
	\tcp*[l]{\footnotesize{Maximum particle distance tensor}}
	$\hat{d}_{\max} \leftarrow \mathbf{MaxDistForParticles}(d_{\max}, \mathbb{P}_t) \in \mathbb{R}^{L_{s, t}}$ \\
	$\tilde{X}_{f, t}$ $\leftarrow \tilde{X}_{t}[\Vert \tilde{X}_{t} - \mathbf{x}_C[: 3] \Vert_{2} - \hat{d}_{\max} \leq \epsilon_H] \in \mathbb{R}^{L_{f, t}}$ \\
	$C(\tilde{Z}_{t}) \leftarrow \mathbf{Denominator}(\tilde{Z}_{t}, \tilde{X}_{f, t})$ \qquad\quad $\triangleright$ Eq. \eqref{c_batch} \\
	$\mathbf{UpdateWeights}(\tilde{Z}_{t}, \tilde{X}_{f, t}, \tilde{X}_{t})$  ~~~~~ $\triangleright$ Eq. \eqref{G_batch}, \eqref{update_tensor}
\end{algorithm}

From \eqref{likelihood}, we firstly accumulate likelihood tensor $\mathrm{g}(\tilde{Z}_t|\tilde{X}_{f, t})$ using the current weight along particle dimension to obtain the denominator $C(\tilde{Z}_t) \in \mathbb{R}^{M_{z, t}}$, with $i$-th element given by
\begin{equation}
	C(\tilde{Z}_t)[i] = P_d \sum_{j = 1}^{L_{f, t}} \left[ W_{f, t|t-1} \odot \mathrm{g}(\tilde{Z}_t|\tilde{X}_{f, t}) \right][i,j] + \sum_{k = 1}^{L_{b, t}} \omega_{b, t}^{(k)} + \kappa,
	\label{c_batch}
\end{equation}
where $W_{f, t|t-1} = (\omega_{s, t|t-1}^{(1)}, ..., \omega_{s, t|t-1}^{(L_{f, t})}) \in \mathbb{R}^{L_{f, t}}$ is a weight tensor of survived and observable particles in FOV, replicating along the first dimension to $\mathbb{R}^{M_{z, t} \times L_{f, t}}$, and $\odot$ denotes element-wise product. In Eq. \eqref{c_batch}, the sum operator in the first term acts on the second dimension $j = 1, ..., L_{f, t}$ and results in $\mathbb{R}^{M_{z, t}}$. $\omega_{b, t}^{(k)}$ denotes the prior weight of newborn particle. The PHD of false detection can be simplified to a constant $\kappa$. From \eqref{g_batch}, we have $\mathrm{g}(\tilde{Z}_t|\tilde{X}_{f, t}) \in \mathbb{R}^{M_{z, t} \times L_{f, t}}$. Following \eqref{likelihood}, \eqref{g_batch} and \eqref{c_batch}, $G(\bm{z}_t, \bm{x})$ in batch style can be calculated by the element-wise division
\begin{equation}
	G(\tilde{Z}_{t}, \tilde{X}_{f, t})[i, j] = \mathrm{g}(\tilde{Z}_{t} | \tilde{X}_{f, t})[i, j] / C(\tilde{Z}_t)[i, j],
	\label{G_batch}
\end{equation}
where $i \in [1, M_{z, t}]$, $j \in [1, L_{f, t}]$ and $i, j \in \mathbb{N}^+$. Here, the second dimension is added to denominator tensor $C(\tilde{Z}_t)$, then we replicate it to shape $\mathbb{R}^{M_{z, t} \times L_{f, t}}$. The normalized likelihood tensor $G(\tilde{X}_{f, t}) \in \mathbb{R}^{L_{f, t}}$ can be obtained by accumulating $M_{z, t}$ observation points, with every element given by $G(\tilde{X}_{f, t})[j] = \sum_{i = 1}^{M_{z, t}} G(\tilde{Z}_{t}, \tilde{X}_{f, t})[i, j]$. Then, it is easy to update weights for survived and observable particles as 
\begin{equation}
	W_{f, t} = \left( 1 - P_d + P_d G(\tilde{X}_{f, t}) \right) \odot W_{f, t|t-1},
	\label{update_tensor}
\end{equation}
where $\odot$ is the element-wise product between two tensors. The updated weight tensor is denoted by $W_{f, t} = (\omega_{t}^{(1)}, ..., \omega_{t}^{(L_{f, t})})$. To handle the occlusion problem, we only update weights for observable particles before estimated thickness of occlusion. For invisible particles, their weights and positions are maintained. The weight update process is shown in Alg. \ref{alg2}.

Through the tensor operation for our map, there are several remarkable advantages. The most significant strength is that our method greatly reduces the time complexity in SMC-PHD filter. The time complexity of the standard DSP \cite{chen2024continuous} exceeds $O(\frac{\theta^2}{\theta_h\theta_v}L_{z,t} L_{f,t})$. Due to the large number of valid particles and observed points, this will slow down the mapping process. However, it is necessary to guarantee real-time performance in reactive planning for manipulators. Low frequency mapping tends to lose track of some moving obstacles in manipulation, making it difficult to react to them. Instead of traversing each particle or observation sequentially on CPU, batch operations gives the possibility for massively parallel computation with the idea of trading GPU space for time. The time complexity of weight update using batch operation is nearly $O(1)$, which also provides more accurate weight updates for particles.

\subsection{Trilevel Model and Connection With Planning}

\begin{figure}
	\centering
	\setlength{\abovecaptionskip}{0cm}
	\includegraphics[width=1.0\linewidth]{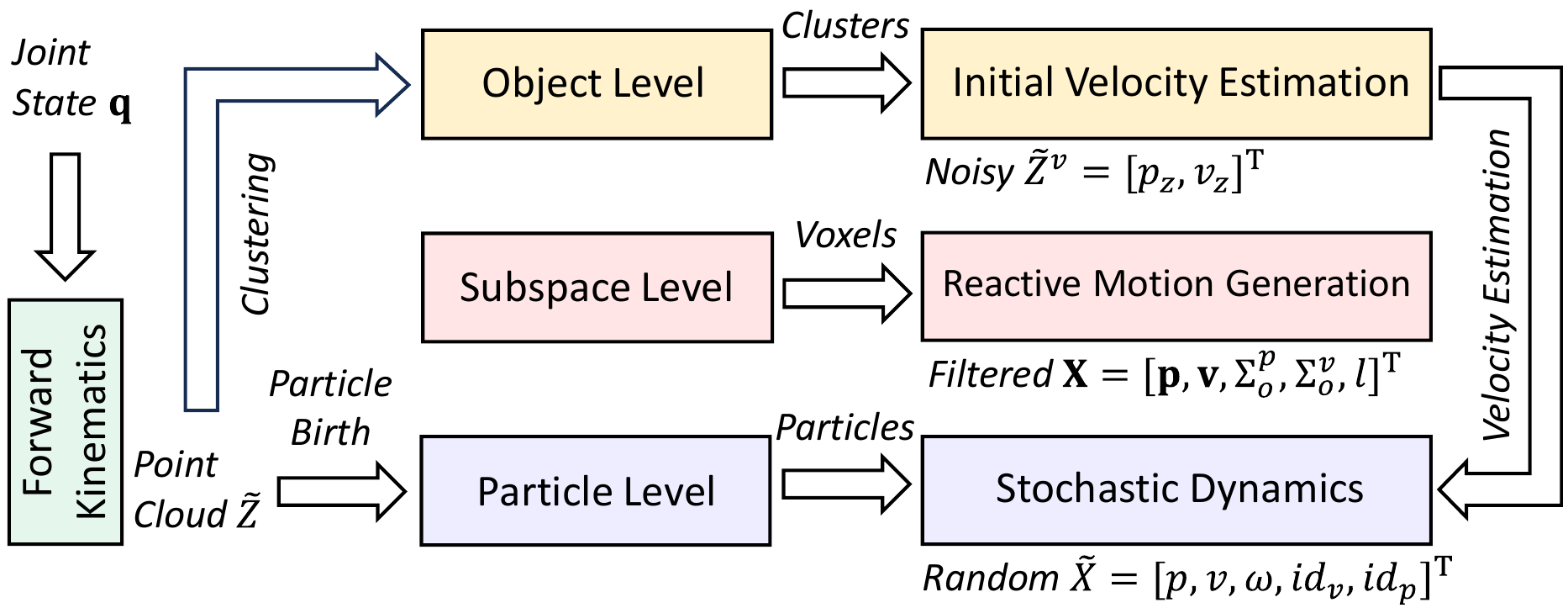}
	\caption{Trilevel model of the G-DSP map. The mapping process of particle-based maps can be divided into three levels, including object (cluster) level, subspace level and particle level. The object level utilizes the finite difference of cluster center positions at time step $t + 1$ and $t$ to estimate rough velocities. The pyramid and voxel subspaces play the significant role in particle weight update and occupancy estimation. The core of G-DSP map is the stochastic dynamics embedded to particle representation, which is realized by parallel tensor operation. For planning, voxels with velocity and uncertainty estimation are considered as obstacle primitives.}
	\label{fig:TrilevelModel}
\end{figure}

The efficient global mapping for limited manipulation environments employs an unique trilevel model, see Fig. \ref{fig:TrilevelModel}. With the coordination among three levels using SMC-PHD filter, particle-based maps achieve estimations to both positions and velocities of voxels. Specifically, stochastic dynamics has been introduced for mapping through particle-based representation for the point objects. Notably, the difference between G-DSP and DSP is three-fold. First, the velocities and uncertainties of voxel primitives are explicitly considered, reducing primitive randomness and computational overhead for planner. Second, batch-operated fashion effectively accelerates propagation and newborn for particles, making the update of weights more accurate through tensor optimizations. For the practical deliberation, the kinematics and geometry of articulated manipulator are considered to avoid possible mistake of the observed point cloud of its own bodies. 

The existing problem remains how to express results of G-DSP map for reactive planning in a reliable and efficient way. As discussed in Section \ref{Sec:RelatedWork}, simple primitives are generally used to model arbitrary-shaped obstacles. Similarly, the articulated bodies of the multi-DOF manipulator are also approximated by simple geometries, such as spheres \cite{zucker2013chomp, bhardwaj2022storm}, tapered capsules \cite{michaux2024safe} and SDFs \cite{koptev2024reactive} etc. To balance accuracy and speed of robot representation, we utilize Bernstein polynomial basis in \cite{li2024representing} to calculate SDFs for each link of manipulator. For such case, it is expensive to obtain distances between SDFs and SDFs, or other complex computational geometries. Different from those static objects, neither particles nor clusters is accurate enough. Thus, the simple voxel primitives are ideal for manipulators. For all $N_o$ occupied voxels in time step $t$, we concatenate the center position $\mathbf{p} \left(\mathbb{V}_k\right)$, velocity $\mathbf{v} \left(\mathbb{V}_k\right)$, covariance $\Sigma_o^\mathrm{p} \left(\mathbb{V}_k\right)$, $\Sigma_o^\mathrm{v} \left(\mathbb{V}_k\right)$ and voxel length $l$ to a tensor $\mathbf{X}_t$ at each time step.

\begin{algorithm}[t]
	\label{alg3}
	\caption{Each Step in G-DSP}
	\SetAlgoLined
	\DontPrintSemicolon
	\KwIn{Joint Position $\mathbf{q}_{t}$, Point Cloud $Z_t$, $\mathrm{FK}(\cdot)$, SDF, Occupied Threshold $\epsilon_{occ}$, Particle Tensor $\tilde{X}_{t-1}$
	}
	\KwOut{Tensor of Occupied Position, Velocity and Covariance $\mathbf{X}_t = (\mathbf{p}, \mathbf{v}, \Sigma_o^\mathrm{p}, \Sigma_o^\mathrm{v}, l) \in \mathbb{R}^{N_o \times 25}$}
	$\mathbf{x}_C, \mathbf{T}_{wc} \leftarrow \mathbf{GetCameraPose}()$ \\
	$\tilde{Z}_t \leftarrow \mathbf{VoxelFilterAndTransform}(Z_t, \mathbf{T}_{wc})$ \\
	$\mathbf{RemoveInsidePoints}(\tilde{Z}_t, \partial\mathcal{R}(\mathbf{q}_t))$ 
	\\ \tcp*[l]{\scriptsize{A child thread to estimate velocity for clusters}}
	$\psi \leftarrow \mathbf{VelocityEstimationThread}(\tilde{Z}_{t}, \tilde{Z}_{t-1})$ \\
	$\tilde{X}_{t} \leftarrow \mathbf{Predict}(\tilde{X}_{t-1})$ \qquad\qquad~\qquad~\qquad $\triangleright$ Eq. \eqref{predict} \\
	$\tilde{X}_{t} \leftarrow \mathbf{Update}(\tilde{X}_{t}, \tilde{Z}_{t}, \mathbf{x}_C)$ \qquad~~~\qquad~\qquad $\triangleright$ Alg. \ref{alg2} \\
	\tcp*[l]{\scriptsize{Wait until the child thread is finished}}
	$\tilde{Z}^v_t \leftarrow \psi.\mathbf{join}()$ \tcp*[r]{\scriptsize{$\tilde{Z}^v_t$ is point tensor with velocity}}
	$\mathbb{V}_t \leftarrow \mathbf{AssignParticlesToVoxels}(\tilde{X}_{t})$ \\
	$P_{occ}\left(\mathbb{V}_t\right) \leftarrow \mathbf{EstimateOcc}(\mathbb{V}_t)$ \qquad\qquad~ $\triangleright$ Eq. \eqref{occupancy_estimation} \\
	$\mathbf{p}\left(\mathbb{V}_t\right) \leftarrow \mathbf{EstimatePos}(\mathbb{V}_t)$ \\
	$\mathbf{v}\left(\mathbb{V}_t\right) \leftarrow \mathbf{EstimateVel}(\mathbb{V}_t)$ \qquad\qquad~~~~~ $\triangleright$ Eq. \eqref{velocity_estimation} \\
	$\Sigma_o^\mathrm{p}\left(\mathbb{V}_t\right), \Sigma_o^\mathrm{v}\left(\mathbb{V}_t\right) \leftarrow \mathbf{EstimateCov}(\mathbb{V}_t)$ ~~~ $\triangleright$ Eq. \eqref{covariance_estimation} \\
	$\tilde{X}_{t} \leftarrow \mathbf{Resample}(\mathbb{V}_t)$ \tcp*[r]{\scriptsize{SIR}}
	$\tilde{X}_t \leftarrow \mathbf{AddNewbornParticles}(\tilde{X}_t, \tilde{Z}^v_t)$ \\
	$\mathbf{X}_t \leftarrow \mathbf{GetOccupancyStatus}(\epsilon_{occ}, \mathbb{V}_t)$
\end{algorithm}

\subsection{G-DSP Map}

The general pipeline of G-DSP map is detailed in Alg. \ref{alg3}. For initial velocity estimation for point cloud (Line 4), we use the finite difference between two adjacent time steps in the cluster level. To extract clusters from filtered point clouds, we apply Euclidean distances to construct KD-tree for segmentation and use Munkres algorithm for matching. The distances between all paired centers of clusters are computed as a cost matrix. If there is a cluster not matched, we consider it as a newborn obstacle and temporarily assign its points the zero velocity. For particle birth, $L_{b, t}$ newborn particles $\tilde{X}_{b, t}$ with each weighted by $\sum_{i = 1}^{M_{z, t}} \frac{\omega_{b, t}^{(k)}}{C(\tilde{Z}_t)[i]}$ at observation points are added to particle tensor $\tilde{X}_t$. In resampling step (Line 13), Sequential Importance Resampling (SIR) \cite{arulampalam2002a} is used to prevent particle degeneracy. The particles with higher weights in voxels are more likely to survive or be replicated. We realize SIR using batch operation for all occupied voxels to decrease time consumption. Instead of employing the costly voxel-wise weighted sampling, $\frac{L_{\max}}{N_v}$ random numbers are sampled, which is used to query a certain normalized weight interval corresponding to $\frac{L_{\max}}{N_v}$ particles.

For analysis of computational complexity, the main thread of our G-DSP utilizes tensors to optimize the time complexity. Notably, 
the time complexity of particle prediction, assignment to subspaces, particle birth, weight update, voxel estimation and resampling are all $O(1)$ using tensor optimization in our map. Thus, the most computationally intensive part is the sort operation for particle tensor before assigning to dual subspaces theoretically, which is quasi-linear to the maximum number of particles, i.e. $O(L_{\max})$. Compared to $O(\frac{\theta^2}{\theta_h\theta_v}L_{\max} L_{z, t})$ of DSP, the batch operation realizes the substantial speedup for SMC-PHD filer.
For the child thread, we employed density-based spatial clustering of Open3D \cite{Zhou2018} to generate clusters, resulting to the time cost of nearly $O(L_{z, t} \log L_{z, t})$. Although Munkres algorithm runs in time of $O(N_{c}^3)$, the few number of clusters $N_c$ in many manipulation tasks make it negligible. Theoretically, the algorithm complexity for G-DSP will reach $O(L_{\max})$ at the worst case, which is less than the overall time complexity $O(\frac{\theta^2}{\theta_h\theta_v}L_{\max} L_{z, t})$ of standard DSP.

\section{Reactive Collision-Free Motion Generation in Dynamic Environments}
\label{Sec:Approach2}

In this section, we formulate the general reactive planning in dynamic environments as a stochastic optimal control problem. Then the path integral theory MPPI and STORM are utilized to develop a Dynamic Obstacle-Aware STORM (D-STORM) with parallel tensor optimization in joint space.

\subsection{Simulation Using Batch Operation}
\label{SubSec:Simulation}

Suppose there are $N_o$ obstacle primitives (occupied voxels) with states (i.e. positions, velocities, covariance and resolution) $\mathbf{x}^{(i)} = (\mathbf{p}^{(i)}, \mathbf{v}^{(i)}, \Sigma_o^{\mathrm{p}(i)}, \Sigma_o^{\mathrm{v}(i)}, l) \in \mathbb{R}^{25}$, $i = 1, ..., N_o$, from perception in G-DSP map. To realize the reactive planning for a $d$-DOF manipulator, we utilize an approximated dynamical model $\tilde{\bm{f}} = (\tilde{f}_0, \tilde{f}_1, ..., \tilde{f}_{H-1})$ for simulation within a variable time horizon $H: \mathbf{dt} = (\mathrm{dt}_{0}, \mathrm{dt}_{1}, ..., \mathrm{dt}_{H - 1})$. As introduced in Section \ref{storm}, our objective is to optimize a control sequence $U = (\mathbf{u}_0, \mathbf{u}_1, ..., \mathbf{u}_{H-1})$ to generate a feasible collision-free trajectory in dynamic environments. The control problem can be formulated as
\begin{equation}
	U^{*} = \underset{U~ \in ~\mathcal{U}}{\arg\min} ~ J(U),
\end{equation}
with the objective $J(U)$ is expressed by
\begin{align}
	\mathbb{E}& \left[\sum_{h = 0}^{H - 1} \gamma^{h} \left( c\left(\mathbf{q}_{h}, \bigcup_{i = 1}^{N_o} \mathbf{x}_{h}^{(i)}\right) + \frac{1}{2} \mathbf{u}_{h}^\top R \left(\mathbf{u}_{h} + 2 \delta \mathbf{u}_{h}\right) \right) \right], \\
	\mathrm{s.t.} ~ &\mathbf{q}_{h+1} = \tilde{f}_{h} \left( \mathbf{q}_{h}, \mathbf{w}_{h} \right), \delta\mathbf{u}_{h} \sim \mathcal{N} \left( \mathbf{0}, \Sigma_{c} \right), \\
	&\mathbf{q}_{0} = \mathbf{q}_{t}, \mathbf{x}^{(i)}_{0} = \mathbf{x}^{(i)}_{t}, i = 1, ..., N_o,
\end{align}
where $R \in \mathbb{R}^{d \times d}$ is a positive definite control matrix, while the cost function $c(\tau)$ consists of two parts: a typical control cost and a state-dependent immediate cost $c\left(\mathbf{q}_{h}, \bigcup_{i = 1}^{N_o} \mathbf{x}_{h}^{(i)}\right)$, providing the implicit constraint to obstacle avoidance. $\gamma \in (0, 1]$ is the discount factor. Note that we explicitly consider the cost for predicted robot and obstacle states. The states of robot and the $i$-th obstacle in the time step $h$ are denoted by $\mathbf{q}_{h}$ and $\mathbf{x}^{(i)}_h$ respectively. Once the costs of all $K$ sampled trajectories are computed, the optimal control will be derived from \eqref{storm_equ1}, \eqref{storm_equ2}, \eqref{storm_equ3}. Thus, the primary work is to determine states along MPPI horizon and design immediate costs in dynamic environments.

For manipulator, an approximated dynamical model $\tilde{\bm{f}} $\cite{bhardwaj2022storm} is used to obtain batched joint position sequences $\Theta \in \mathbb{R}^{K \times H \times d}$ in parallel given sampled control sequences $\{V_k\}_{k = 1}^{K}$. For $\Theta$, every element $\mathbf{q}_{h, k} \in \mathbb{R}^d$ denotes robot joint position at time step $h$ in the $k$-th trajectory. Let us consider $V = \{V_k\}_{k = 1}^{K} \in \mathbb{R}^{K \times H \times d}$. For variable time step $\mathbf{dt} = (\mathrm{dt}_{0}, \mathrm{dt}_{1}, ..., \mathrm{dt}_{H - 1}) \in \mathbb{R}^{H}$ along horizon, we have
\begin{align}
	\ddot{\Theta} &= V, \notag \\
	\dot{\Theta} &= \dot{\Theta}_{0} + T_{l}\left(1\right) \mathrm{diag} \left(\mathbf{dt}\right) \ddot{\Theta}, \label{approx_robot_model_equ} \\
	\Theta &= \Theta_{0} + T_{l}\left(1\right) \mathrm{diag} \left(\mathbf{dt}\right) \dot{\Theta}, \notag
\end{align}
where $T_l \left(1\right) \in \mathbb{R}^{H \times H}$ is a lower triangular matrix for the last two dimensions, and $\mathrm{diag} \left(\mathbf{dt}\right) \triangleq \mathrm{diag} \left( \mathrm{dt}_{0}, \mathrm{dt}_{1}, ..., \mathrm{dt}_{H - 1} \right) \in \mathbb{R}^{H \times H}$,  which are then replicated along extended first dimension to $\mathbb{R}^{K \times H \times H}$. Here, $\Theta$, $\dot{\Theta}$, $\ddot{\Theta}$ are position, velocity and acceleration tensor respectively, while $\Theta_{0}, \dot{\Theta}_{0}$ are current joint position and velocity. By multiplying the lower triangular matrix $T_l \left(1\right)$, the accumulation for joint velocities and positions is efficiently implemented along horizon. To balance the length of MPPI lookahead and obstacle impenetrability, we also apply a smaller timestep near current robot state and the larger ones for later steps. Then, the robot joint state tensor of $K$ sampled trajectories along $H$ horizon $(\Theta, \dot{\Theta}, \ddot{\Theta})$ is obtained.

\begin{figure}
	\centering
	\setlength{\abovecaptionskip}{0cm}
	\includegraphics[width=1.0\linewidth]{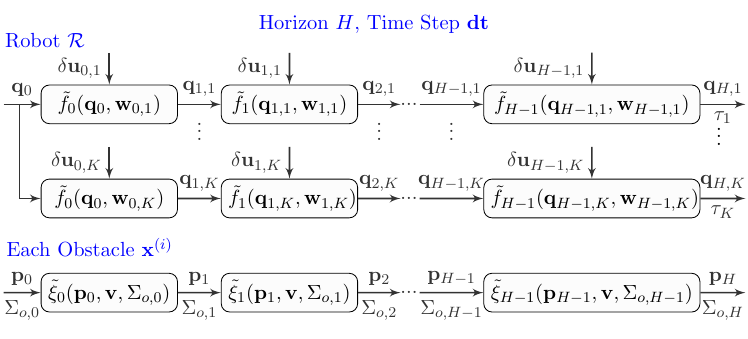}
	\caption{Illustration of the system dynamics propagation in D-STORM. The robot considers uncertainty by sampling $K$ control sequences with importance weighting, while obstacle voxels explicitly propagate their uncertainties by covariance. In D-STORM, the horizon of obstacles are aligned with the robot by time step $\mathbf{dt}$, which is used to rollout in $H$. Then the immediate cost can be calculated by spatial positions of robot-obstacle dynamics in the horizon. }
	\label{fig:D_STORM}
\end{figure}

For obstacles, a constant velocity model is employed based on the parameters of G-DSP map. Let us use 
\begin{equation}
	\mathcal{X} = \left[ \begin{array}{ccc}
		\mathbf{p}_{0}^{(1)} & ... & \mathbf{p}_{H-1}^{(1)} \\
		... & ... & ... \\ 
		\mathbf{p}_{0}^{(N_o)} & ... & \mathbf{p}_{H-1}^{(N_o)}
	\end{array} \right] \in \mathbb{R}^{N_o \times H \times 3}, \notag
\end{equation}
with $\mathbf{p}_{h}^{(i)} \in \mathbb{R}^3$, $i = 1, ..., N_o, h = 0, ..., H - 1$, to denote the simulated position tensor of $N_o$ obstacles along horizon $H$. We also duplicate obstacle velocity tensor $\dot{\mathcal{X}} = (\mathbf{v}^{(1)}, ..., \mathbf{v}^{(N_o)}) \in \mathbb{R}^{N_o \times 3}$ along newly added second horizon dimension to shape $\mathbb{R}^{N_o \times H \times 3}$. Using the same time step $\mathbf{dt}$ as \eqref{approx_robot_model_equ}, the predicted obstacle position can be expressed as
\begin{equation}
	\mathcal{X} = \mathcal{X}_{0} + T_{l}\left(1\right) \mathrm{diag} \left(\mathbf{dt}\right) \dot{\mathcal{X}},
	\label{approx_obstacle_model_equ}
\end{equation}
where $\mathcal{X}_0 = (\mathbf{p}^{(1)}, ..., \mathbf{p}^{(N_o)}) \in \mathbb{R}^{N_o \times 3}$ is the current observed position tensor of obstacles. For alignment, we add the second dimension for $\mathcal{X}_0$ and replicate along the new horizon axis to $\mathcal{X}_0 \in \mathbb{R}^{N_o \times H \times 3}$. 
Due to uncertainty from real observation, the covariance along planning horizon should also be propagated. Thus, we propagate uncertainty $\Sigma_{o}$ as Extended Kalman Filter
\begin{equation}
	\Sigma_{o, h+1} = A_{h} \Sigma_{o, h} A_{h}^\top + Q,
	\label{mppi_obstacle_cov_equ}
\end{equation}
where the transition matrix at time step $h$ can be denoted as $A_{h} = \left[ \begin{array}{cc}
	I_{3 \times 3} & \mathrm{dt}_h I_{3 \times 3} \\ \mathbf{0}_{3 \times 3} & I_{3 \times 3}
\end{array} \right]$, $\Sigma_o = 
\left[ \begin{array}{cc}
	\Sigma_o^\mathrm{p} & \mathbf{0}_{3 \times 3} \\ \mathbf{0}_{3 \times 3} & \Sigma_o^\mathrm{v}
\end{array} \right]$, $\mathrm{dt}_h$ is the $h$-th period of $\mathbf{dt}$, and $Q$ is the additive process noise. To compute all $N_o$ obstacles in batch, we add the first dimension for $A_h$ and extend it to $\mathbb{R}^{N_o \times 6 \times 6}$, transforming $N_o$ obstacle covariance matrix to $\Sigma_{o} \in \mathbb{R}^{N_o \times 6 \times 6}$ from $h$ to $h+1$. The transpose acts on the last two dimensions of expanded tensor $A_h$. From Eq. \eqref{mppi_obstacle_cov_equ}, the uncertainty of obstacle states $\Sigma_{o}$ can be solved sequentially along control horizon. We call Eq. \eqref{approx_obstacle_model_equ} and \eqref{mppi_obstacle_cov_equ} the approximated dynamical model for obstacles $\tilde{\bm{\xi}} \triangleq (\tilde{\xi}_0, \tilde{\xi}_1, ..., \tilde{\xi}_{H-1})$. The rollouts are illustrated in Fig. \ref{fig:D_STORM}. Clearly, the dynamics rollouts not only propagates robot positions, but also handles positions, velocities and uncertainties for dynamic obstacle in the proposed D-STORM.

\subsection{Uncertainty-Aware Cost for Dynamic Obstacles}

To handle complex-shaped obstacles, a complete dynamics of $N_o$ voxel primitives is perceived by G-DSP map. Given the propagated positions of $K$ robot trajectories in joint space and $N_o$ obstacle primitives along horizon $H$, we estimate collision costs between them. After spatial-temporal alignment of robot-obstacle dynamics in Section \ref{SubSec:Simulation}, we utilize an uncertainty-aware soft manner considering the stochastic nature of MPPI.

For deterministic cases, each obstacle primitive is generally modeled as a sphere with fixed radius $r_o = \frac{\sqrt{3}}{2} l + \delta r$, where $\delta r$ is called safe threshold. Suppose the estimated position for the $i$-th voxel is $\mathbf{p}^{(i)}_{h}$ at horizon $h$, and $\mathbf{p}^{(i)}_{h, s} \in \partial \mathcal{R}$ denotes the point on robot surface which is the closest to voxel center $\mathbf{p}^{(i)}_{h}$. Then, the signed distance is defined by $\mathrm{dist}(\mathbf{p}_h^{(i)}) = \pm \Vert \mathbf{p}_h^{(i)} - \mathbf{p}^{(i)}_{h, s} \Vert_{2}$, and the gradient of SDF is $\mathbf{n}_{h}^{(i)} = \nabla_{\mathbf{p}_h^{(i)}} \mathrm{dist}(\mathbf{p}_h^{(i)})$. For widely used spherical approximation, the cost is formulated as
\begin{equation}
	c_{col} (\mathbf{q}_h) = c_{col} \left(\mathbf{q}_h, \bigcup_{i = 1}^{N_o} \mathbf{x}_{h}^{(i)} \right) = \max_{i = 1, ... , N_o} c_{col}(\mathbf{x}^{(i)}_h, \mathbf{q}_h),
	\label{deterministic_cost}
\end{equation}
where
\begin{equation}
	c_{col}(\mathbf{x}^{(i)}_h, \mathbf{q}_h) = \left\{ \begin{array}{ll}
		r_o - \mathrm{dist}(\mathbf{p}_h^{(i)}), &\mbox{if} ~ \mathrm{dist}(\mathbf{p}_h^{(i)}) < r_o, \\ 
		0, &\mbox{otherwise}.
	\end{array} \right.
	\notag
\end{equation}
By considering the nearest obstacle primitive, classical MPPI-based planning methods use a fixed threshold to penalize those trajectories that approach or penetrate obstacle geometry.

However, real observation of camera tends to be uncertain, especially for point cloud in motion. For $i$-th voxel primitive, we consider $\Sigma_{o, h}^{\mathrm{p} (i)} \in \mathbb{R}^{3 \times 3}$ as position covariance at $h$-th time step estimated by the G-DSP map and dynamics propagation. The collision distribution is $\hat{\mathbf{p}} \sim \mathcal{N}(\mathbf{p}^{(i)}_{h}, \Sigma_{o, h}^{\mathrm{p}(i)})$. Here, we also use regularization to ensure the positive definiteness for $\Sigma_{o, h}^{\mathrm{p}(i)}$.

With Gaussian approximation, we define a covariance ellipsoid to explicitly measure uncertainty as
\begin{equation}
	\mathcal{E}_\nu = \{ \hat{\mathbf{p}}: (\hat{\mathbf{p}} - \mathbf{p}_h)^\top \Sigma_{o, h}^{-1} (\hat{\mathbf{p}} - \mathbf{p}_h) \leq \nu^2 \},
\end{equation}
where the covariance $\Sigma_{o, h} \succ \mathbf{0}$, and $\nu^2$ denotes the confidence level that represents chi-square quantile. We omit the obstacle number $i$ for a clearer derivation. Given confidence probability and dimension of distribution,
the radial distance from obstacle ellipsoid to the center along $\mathbf{n}_h$ is given by
\begin{equation}
	r_e(\mathbf{n}_h) = \nu / \sqrt{\mathbf{n}_h^\top \Sigma_{o, h}^{-1} \mathbf{n}_h}.
	\label{ellipsoid_radius}
\end{equation}
For the maximum radial distance given $\mathbf{n}_h$ on convex set $\mathcal{E}_\nu$, the support function can be expressed by
\begin{equation}
	\sigma_{\mathcal{E}_\nu}(\mathbf{n}_h) = \sup_{\hat{\mathbf{p}} \in \mathcal{E}_\nu} \mathbf{n}_h^\top \hat{\mathbf{p}}
	= \nu \sqrt{\mathbf{n}_h^\top \Sigma_{o, h} \mathbf{n}_h} + \mathbf{n}_h^\top \mathbf{p}_h,
	\label{support_function}
\end{equation}
where the first term provides a tight upper bound for the radial distance in Eq. \eqref{ellipsoid_radius}, while the second term is center offset. Intuitively, the radial distance of the obstacle ellipsoid is longer if the uncertainty along that direction is larger. Theoretically, replacing $r_o$ in Eq. \eqref{deterministic_cost} with $\nu \sqrt{\mathbf{n}_h^\top \Sigma_{o, h} \mathbf{n}_h} + \delta r$ can handle the uncertainty of dynamic obstacles. Nonetheless, it is intractable for expensive computation of high-dimensional exact gradient tensors $\mathbf{n}_h$ for $K \times H \times N_o$ robot-obstacle pairs. 

In SMART, the uncertainty-aware radius can be slacked by considering its maximum radial distance ($\delta r = 0$),
\begin{equation}
	r_d = \nu \max_{\Vert \mathbf{n}_h \Vert_{2} = 1} \sqrt{\mathbf{n}_h^\top \Sigma_{o, h}^{\mathrm{p}} \mathbf{n}_h}
	= \nu \sqrt{\lambda_{\max} ( \Sigma_{o, h}^{\mathrm{p}} )},
	\label{radius}
\end{equation}
where $\lambda_{\max} ( \Sigma_{o, h}^{\mathrm{p}} )$ denotes the maximum eigenvalue of $\Sigma_{o, h}^{\mathrm{p}}$. 
In practical, we set interval $[\frac{3}{4}l + \delta r, \frac{5}{4}l + \delta r]$ as a bound for radius considering the particle noise, i.e. $r_o = \min(\max(r_d, \frac{3}{4}l + \delta r), \frac{5}{4}l + \delta r)$. According to Eq. \eqref{deterministic_cost} and Eq. \eqref{radius}, we take the maximum cost for all $N_o$ primitives as collision cost at time step $h$ for the $k$-th simulated trajectory. Given confidence, the collision cost is adaptively controlled by closed form of Eq. \eqref{radius} and Eq. \eqref{deterministic_cost}. When the uncertainty is low, $r_d$ is less than $\frac{\sqrt{3}}{2}l$, and vice verse. In experiments, $\nu$ is set to 2.5 ($p \approx 0.95$).

\subsection{Other Costs}

We encode several other costs, including self-collision cost, joint limits cost, manipulability cost and goal cost. We formulate the total cost as a weighted sum of costs as \cite{rakita2018RelaxedIK, bhardwaj2022storm}.

\textit{1) Self-Collision Avoidance}. To compute the closest distance between all pairs of robot links, we train a neural network to predict the minimum distance given a batch of joint positions $\Theta$. To improve the accuracy, jointNerf \cite{bhardwaj2022storm} is used to compute self-collision distance. For dataset collection, we parse parameters of robot kinematics and collision geometry using FCL \cite{pan2012fcl}. After sampling 500,000 joint configurations uniformly, numerical solutions are calculated to form the training set. The self-collision cost term can be written as
\begin{equation}
	c_{self}(\mathbf{q}_{h}) = -\min(0, \mbox{jointNerf}(\mathbf{q}_{h})).
\end{equation} 
When there is no collision between bodies, the signed distance is positive, and vice verse. Thus, we only penalize penetration cases between robot links.

\textit{2) Joint Limits}. To guarantee that the robot manipulates in required joint range, we penalize the joint position $\mathbf{q}$ when it is beyond a given safety threshold $[\mathbf{q}_{\min}, \mathbf{q}_{\max}]$,
\begin{align}
	c_{lim}(\mathbf{q}_{h}) = \left\{
	\begin{array}{ll}
		\Vert \mathbf{q}_{\min} - \mathbf{q}_{h} \Vert_{2}, &\mbox{if} ~ \mathbf{q}_h < \mathbf{q}_{\min}, \\
		\Vert \mathbf{q}_{h} - \mathbf{q}_{\max} \Vert_{2}, &\mbox{else if} ~ \mathbf{q}_{h} > \mathbf{q}_{\max}, \\
		0, &\mbox{otherwise},
	\end{array}
	\right.
\end{align}
where we also reduce joint range slightly and use soft threshold for joint violation penalty to further strength the limits.

\textit{3) Manipulability}. The robotic velocity manipulability measures the ability in generating any arbitrary velocity at given joint position, which is denoted by $m = \sqrt{\det \left(\mathbf{J}(\mathbf{q}) \mathbf{J}(\mathbf{q})^{\top}\right)}$ \cite{yoshikawa1985manipulability, jaquier2021geometry}. We apply the manipulability cost to penalize those configurations with small manipulability index:
\begin{equation}
	c_{manip}(\mathbf{q}_h) = \left\{ \begin{array}{lcl}
		1 - m / \epsilon_{m}, & &\mbox{if} ~ m < \epsilon_{m}, \\ 
		0,& &\mbox{otherwise},
	\end{array} \right.
\end{equation}
where $\epsilon_{m}$ is the positive manipulability threshold, and we use $0.05$ in experiments. With the cost term, the robot can avoid singular configurations in predictive horizon.

\textit{4) Goal Reaching}. Given a goal configuration and end joint configuration of rollouts $\mathbf{q}_{g}, \mathbf{q}_H \in \mathbb{R}^d$, the cost is defined by
\begin{equation}
	c_{goal}(\mathbf{q}_g, \mathbf{q}_H) =  \Vert \mathbf{q}_g - \mathbf{q}_H \Vert_{2},
\end{equation}
where $c_{goal}(\mathbf{q}_g, \mathbf{q}_H)$ allows us to drive robots approach target configuration while avoiding obstacles.

Thus, the overall immediate cost function is given by 
\begin{equation}
	c = \rho_g c_{goal} + \rho_c (c_{col} + c_{self}) + \rho_l c_{lim} + \rho_m c_{manip},
\end{equation}
where the weights $\rho_g, \rho_c, \rho_l, \rho_m$ are defined according to the task considering more important one with a larger value.

\subsection{Practical Discussions}
In this section, we discuss two important issues of practical deployment for the proposed D-STORM, highlighting feasibility of parallelization and tight relevance with G-DSP map.

\begin{algorithm}[t]
	\label{alg4}
	\caption{D-STORM Open-Loop Control}
	\SetAlgoLined
	\DontPrintSemicolon
	\KwIn{Number of Trajectories $K$, Horizon $H$, $\mathbf{dt}$ Robot Dynamics $\tilde{\bm{f}}$, Obstacle Dynamics $\tilde{\bm{\xi}}$ \qquad
		Start/Goal States $\mathbf{q}_{0}, \mathbf{q}_{T}$, Initial Control $U$
		Obstacle State $\mathbf{X} = {(\mathbf{p}, \mathbf{v}, \Sigma_o^\mathrm{p}, \Sigma_o^\mathrm{v}, l)} \in \mathbb{R}^{N_o \times 25}$
	}
	\KwOut{The optimal control $\mathbf{u}_0 = \ddot{\mathbf{q}}_d$}
	$\mathbf{q} \leftarrow \mathbf{q}_{0}$ \\
	\While{task not completed}{
		$\mathbf{q}, \dot{\mathbf{q}}, \ddot{\mathbf{q}} \leftarrow \mathbf{GetRobotStateEstimation}()$ \\
		$\mathbf{X} \leftarrow \mathbf{GetObstacleObservation}()$ \qquad $\triangleright$ Alg. \ref{alg3} \\
		Sample $\delta U = (\delta \mathbf{u}_h^k) \in \mathbb{R}^{K \times H \times d}, \delta \mathbf{u}_h^k \in \mathcal{N}(0, \Sigma_{c})$ \\
		$V \leftarrow U + \delta U$ \\
		\tcp*[l]{\footnotesize{Propagate robot states in parallel}}
		$\Theta, \dot{\Theta}, \ddot{\Theta} \leftarrow \tilde{\bm{f}}(\mathbf{q}, \dot{\mathbf{q}}, \ddot{\mathbf{q}}, V)$ \qquad~\qquad\qquad $\triangleright$ Eq. \eqref{approx_robot_model_equ} \\
		$\mathbf{X}^c \leftarrow \mathbf{TopN_{o}^{c}ClosestObstacles} (\mathbf{X})$\\
		\tcp*[l]{\footnotesize{Propagate obstacle states in parallel}}
		$\mathcal{X}, \Sigma_o \leftarrow \tilde{\bm{\xi}}(\mathbf{X}^c)$ \qquad\qquad\qquad\quad~ $\triangleright$ Eq. \eqref{approx_obstacle_model_equ}, \eqref{approx_obstacle_cov_equ} \\
		\tcp*[l]{\footnotesize{$C = (c(\tau_1), ..., c(\tau_K)) \in \mathbb{R}^{K}, c(\Theta, \mathcal{X}, \Sigma_o) \in \mathbb{R}^{K \times H}$}}
		$C \leftarrow \sum_{h = 0}^{H-1}\gamma^h c(\Theta, \mathcal{X}, \Sigma_o)[:, h]$ \\
		\tcp*[l]{\footnotesize{$\bm{\eta} = (\eta_1, ..., \eta_K) \in \mathbb{R}^{K}$}}
		$\bm{\eta} \leftarrow \mathbf{GetWeights}(C)$ \qquad\qquad\qquad~~ $\triangleright$ Eq. \eqref{storm_equ1} \\
		$U \leftarrow \mathbf{UpdateControl}(U, \bm{\eta})$ ~~~~\qquad\quad $\triangleright$ Eq. \eqref{storm_equ2} \\
		$\Sigma_{c} \leftarrow \mathbf{UpdateCovariance}(\Sigma_{c}, \bm{\eta})$ ~~\quad $\triangleright$ Eq. \eqref{storm_equ3} \\
		$\dot{\mathbf{q}}_d \leftarrow \dot{\mathbf{q}}_d + \mathbf{u}_0 \Delta \tau$, 
		$\mathbf{q}_d \leftarrow \mathbf{q}_d + \dot{\mathbf{q}}_d \Delta \tau$ \\
		$\mathbf{SendToLowLevelController}(\mathbf{q}_d, \dot{\mathbf{q}}_d)$ \\
		\For{$h = 1, ..., H-1$}{
			$\mathbf{u}_{h-1} \leftarrow \mathbf{u}_h$
		}
		$\mathbf{u}_{H-1} \leftarrow \mathbf{Initialize}(\mathbf{u}_{H-1})$
	}
\end{algorithm}

\textit{1) Uncertainty Propagation for Trajectories of Obstacles.} One of the most elegant advantages for STORM is to leverage efficient parallel tensor representation for dynamical rollouts. As simulations in horizon $H$ using \eqref{approx_robot_model_equ}, \eqref{approx_obstacle_model_equ}, batch operation greatly accelerates the propagation process of positions, velocities and accelerations. Different from the static environments, it is necessary to consider uncertainties of predicted motions with the introduction of dynamic obstacles. In Eq. \eqref{mppi_obstacle_cov_equ}, we also propagate covariance $\Sigma_{o}$ for all $N_o$ obstacles sequentially along the horizon $H$. However, such recursive propagation for covariance matrix $\Sigma_{o, h}$ will greatly reduce the simulation rate, limiting its usage for a long horizon. Despite the theoretically feasibility, a more practical formula should be derived.

Let us consider a special case where the initial covariance of obstacle state can be represented by block diagonal matrix, i.e.
$\Sigma_{o, 0} = \left[ \begin{array}{cc}
		\Sigma_{o, 0}^\mathrm{p} & \mathbf{0}_{3 \times 3} \\ \mathbf{0}_{3 \times 3} & \Sigma_{o}^\mathrm{v}
\end{array} \right]$, which is consistent with perception result of our map. For reactive planning with short latency, the process noise $Q$ could be negligible. 
From \eqref{mppi_obstacle_cov_equ}, the analytical formula of the covariance at time step $h$ is
\begin{equation}
	\Sigma_{o, h} = \left[ \begin{array}{cc}
		\Sigma_{o, 0}^\mathrm{p} + \left( \sum_{j = 0}^{h - 1} \mathrm{dt}_j \right)^2 \Sigma_{o}^\mathrm{v} & \left( \sum_{j = 0}^{h - 1} \mathrm{dt}_j \right) \Sigma_{o}^\mathrm{v} \\ \left( \sum_{j = 0}^{h - 1} \mathrm{dt}_j \right) \Sigma_{o}^\mathrm{v} & \Sigma_{o}^\mathrm{v}
	\end{array} \right]. \notag
\end{equation}
Accordingly, it is obvious to propagate the position uncertainty $\Sigma_{o}^\mathrm{p} \in \mathbb{R}^{N_o \times H \times 3 \times 3}$ using batch operation along horizon $H$
\begin{equation}
	\Sigma_{o}^\mathrm{p} = \Sigma_{o, 0}^\mathrm{p} +  (T_{l} \left( 1 \right) \mathbf{dt}) \odot (T_{l} \left( 1 \right) \mathbf{dt}) \odot \Sigma_{o}^\mathrm{v},
	\label{approx_obstacle_cov_equ}
\end{equation}
where $T_l \left(1\right) \in \mathbb{R}^{H \times H}$ is a lower triangular matrix filled with 1, $T_l \left(1\right) \mathbf{dt}$ is used to accumulate horizon from $\mathrm{dt}_0$ to $\mathrm{dt}_{h - 1}$. After squaring $T_l \left(1\right) \mathbf{dt}$, we expand its dimension from $\mathbb{R}^{H}$ to $\mathbb{R}^{N_o \times H \times 3 \times 3}$. Meanwhile, the horizon axis are also added to the initial position covariance $\Sigma_{o, 0}^\mathrm{p}$ and velocity covariance $\Sigma_{o}^\mathrm{v}$, extending them to $\mathbb{R}^{N_o \times H \times 3 \times 3}$. With \eqref{approx_obstacle_model_equ} and \eqref{approx_obstacle_cov_equ}, the state and covariance of obstacles can be calculated using batch operation. Notably, the simplification ensures realtime performance, allowing efficient simulation for dynamic obstacles.

\textit{2) Algorithm Implementation.} Now we outline the proposed D-STORM algorithm in Alg. \ref{alg4}. Since our algorithm leverages parallel execution for multiple sampled trajectories in planning horizon, it is suitable for GPU implementation. Given control parameters, cost functions and approximated system dynamics, D-STORM iteratively updates nominal control and covariance until the task is completed. For each control loop interval $\Delta \tau$, D-STORM propagates trajectories via robot-obstacle dynamics (Line 7-9). Clearly, the planning frequency exhibits clear sensitivity to the number of primitives. This effect arises mainly from the computational characteristics of robot collision model \cite{li2024representing} employed in our approach. Thus, we restrict the collision evaluation to the top 20 closest occupied voxels. The costs of aligned spatial states between $K$ trajectories and $N_o^c$ obstacles ($N_o^c \leq 20$) along control horizon $H$ are evaluated using cost functions (Line 10). To mitigate control latency, multiprocessing is utilized to run D-STORM, streaming desired control to low-level controller at around 50 Hz. For a robot with six-axis force/torque sensor, an admittance controller is deployed. We use Robot Operating System ($\mathrm{ROS}$) to communicate between perception and control nodes (Line 3-4).

\begin{figure*}
	\centering
	\setlength{\abovecaptionskip}{-0.10cm}
	\includegraphics[width=1.0\textwidth]{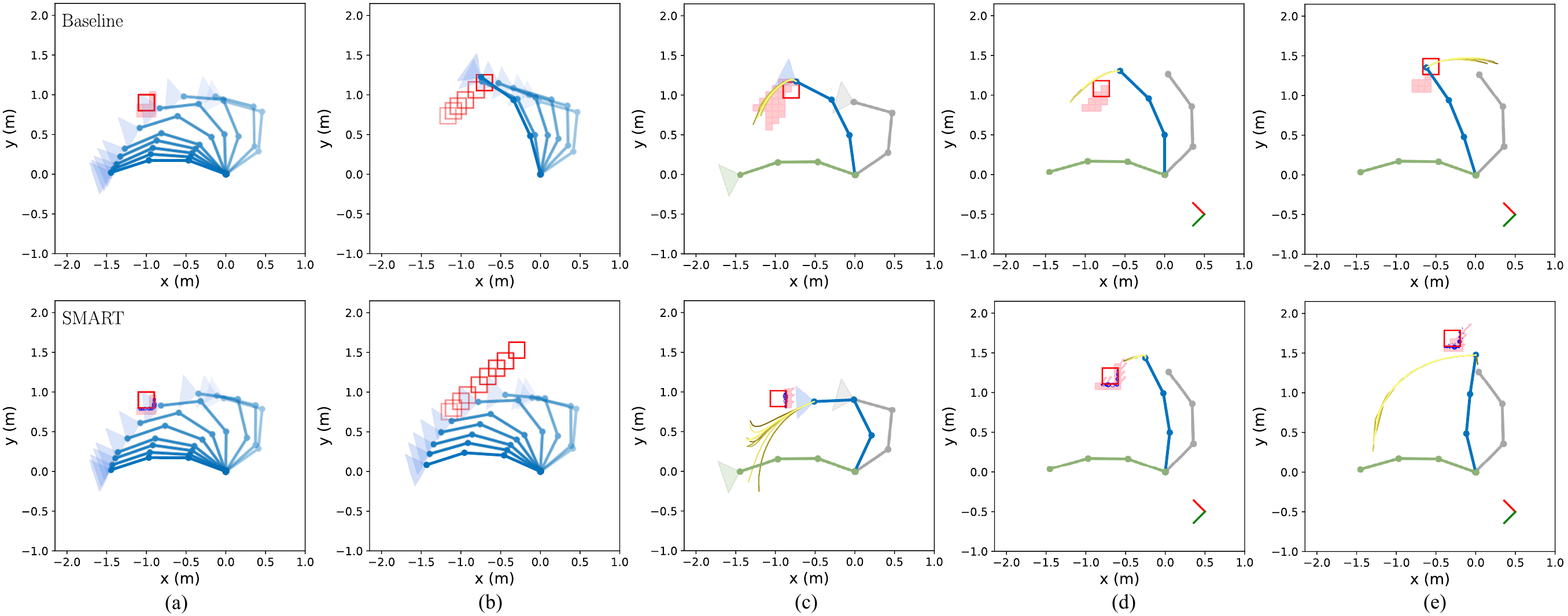}
	\caption{The motions generated from the baseline (top) and SMART (bottom) in the task space. (a)-(c) show cases of robot planning with an onboard sensor, while (d)-(e) visualize a case with a fixed camera. In (a), a 3-DOF manipulator can avoid the static obstacle using both baseline and SMART. However, for fast-moving obstacles in (b) and (d), STORM controller with raycast-based perception loses to capture their dynamic characteristics, leading to a local minima and collision. SMART leverages more complete dynamics estimation and prediction, realizing repulsive reaction to dynamic environments. (c), (d) and (e) details the map and top ten control trajectories in MPC controller at a certain moment. A 3-DOF planar robot (blue) traverses the task space avoiding the red obstacle from initial configuration (gray) to the target configuration (green), modeling the surroundings with voxels (light red grids). 
	}
	\label{fig:NumericalCase}
\end{figure*}

\textit{3) Tight Correlation With Particle-Based Mapping.} To avoid dynamic obstacles in manipulations, the position, velocity and uncertainty estimations for observed obstacles are important. In real-world unstructured cases, there exist arbitrarily shaped obstacles. Many existing perception methods for manipulators model surroundings with primitives like point clouds, voxels or SDFs \cite{ren2023robot, kappler2018real, bhardwaj2022storm, usenko2017real, zucker2013chomp}, which are used to solve the optimal control by only propagating fixed obstacle positions along planning horizon. Even for realtime perception and planning, there is still a risk of collision with dynamic obstacles. Notably, particle-based maps provide estimations for position, valuable velocity and uncertainty by embedding the stochastic dynamics of point objects into SMC-PHD filter. Now, a 3-DOF planar manipulator in dynamic and unstructured scenarios will illustrate the advantage of SMART in sensor-based planning.

\textbf{Numerical Example.} (3-DOF Case) Real-time perception with particles provides position, velocity and uncertainty estimation of obstacles, which is important for reactive motion generation. The MPPI-based control also facilitates real-time acquisition of the latest perception for dynamic obstacles. In this numerical case, we apply different mapping and planning approaches given a 3-DOF planar serial manipulator to reach the goal configuration while observing the surroundings. The conventional framework simulates robot dynamics with fixed obstacle positions, attempting to avoid obstacles. To intuitively compare performances between the proposed framework and the baseline, we consider two schemes: (a) the commonly used raycast-based map \cite{usenko2017real, kappler2018real} is deployed to generate occupied voxels, which are fed into the STORM controller as obstacle primitives (as baseline); (b) the proposed SMART via G-DSP map and D-STORM. Refer to Fig. \ref{fig:NumericalCase} for a 2-D visualization. 

Consider that a 3-DOF planar robot traverses 2-D task space avoiding a red obstacle from the initial to target configuration, modeling the surroundings with voxels. For particle-based map G-DSP, blue hollow circles visualize the particles, where the larger one represents a greater weight. In static environments, both raycast-based and particle-based perception have similar performance. In contrast, the raycast-based map faces several limitation due to accumulated probabilistic inertia and limited FOV. Firstly, voxels for moving obstacles tend to be wrongly estimated for trail noise or occlusion. Also, STORM conservatively optimizes control term using fixed obstacle positions, making it difficult to avoid fast-moving obstacles. Thanks to the inclusion of robot-obstacle dynamics, SMART realizes the reflective obstacle avoidance with dynamic prediction.

\section{Experiments And Evaluations}
\label{Sec:Experiments}

This section evaluates the proposed G-DSP map, D-STORM and SMART by deploying them on a simulated and real UR5 robot to quantify performances, benchmarking against existing state-of-the-art realtime mapping and planning strategies. 

\subsection{Experimental Setup}
\label{SubSec:ExpSetup}
In simulation experiments, we deploy the proposed realtime perception and planning algorithm on a 6-DOF UR5 manipulator with a fixed depth camera Intel RealSense d435i. 
For quantitative evaluations, 
$\rm{PyBullet}$ simulator is used to build several different dynamic scenarios. A desktop equipped with an Intel i7-13700KF CPU and a NVIDIA RTX 4070 GPU is used. Real-world tests are deployed on the real UR5 platform and two fixed RealSense cameras. 
We implement our sensor-based planning framework using $\rm{Python}$ and $\rm{PyTorch}$, which supports CUDA accelerations. Subsequent experiments from Sec. \ref{exp1} to \ref{exp3} are designed to validate three hypotheses:
\begin{itemize}
	\item[1)] Batch operations using tensor representations effectively accelerate particle-based maps to realtime level, particular for perception in manipulation tasks.
	\item[2)] The explicit propagation for dynamic obstacles improves the performance of collision avoidance for arm in MPPI-based manipulations.
	\item[3)] SMART framework has unique advantage in dynamic and real-world environments with complex-shaped obstacles.
\end{itemize}
Only a fixed RGB-D camera is mounted for mapping comparisons.
In real-world sensor-based planning, collision avoidance tasks are designed following \textit{trivially unstructured Assumption}, 
where two depth cameras are fixed to broaden FOV.
The map volumes of evaluations are determined by tasks.

\begin{figure}
	\centering
	\setlength{\abovecaptionskip}{0cm}
	\includegraphics[width=0.95\linewidth]{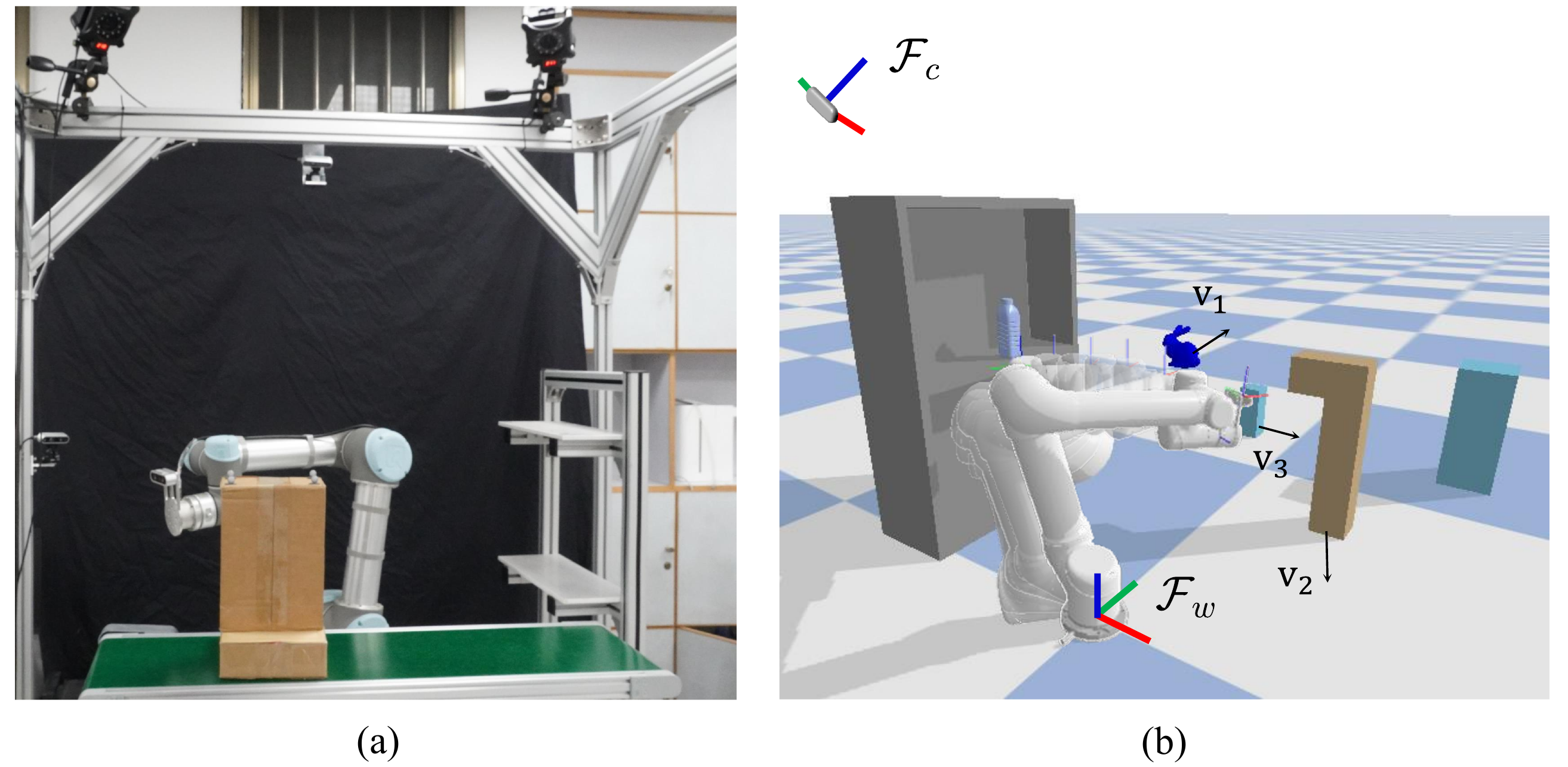}
	\caption{The real-world and simulation environments for mapping performance comparison. (a) The labeled object is moving in the fixed FOV. We consider the synchronized velocity from the NOKOV motion capture system as ground truth. (b) For the simulated occupancy estimation scenario, multiple models are placed on cabinet or the table. The manipulator moves along a predefined trajectory between the initial position and target configuration back and forth. Since the mapping performance is compared here, we ensure that there is no collision between obstacles and the manipulator.}
	\label{fig:MapEnv}
\end{figure}

\subsection{Mapping and Perception Comparison}
\label{exp1}

For perception in manipulations, three important metrics are chosen to quantify dynamic mapping performance, i.e. velocity estimation, binary classification and the computation time. We compared G-DSP map with three mainstream maps, including a raycast-based map Ewok \cite{usenko2017real}, a generalized counting sensor model-based dynamic map K3DOM \cite{min2021kernel} and a SOTA particle-based map DSP \cite{chen2024continuous}. To enable a fair comparison, these maps are modified to the global version and directly receive camera pose. For the velocity estimation, the naive finite difference of two matched cluster centers by Munkres method and DSP are considered as baselines. These maps span $[2.4 \times 2.4 \times 1.5] \mathrm{m}$.

To evaluate the velocity estimation performance of map for manipulators, an Intel RealSense is mounted on a fixed support located behind and above the robot to collect point cloud. The filtered points are used to generate the baseline Munkres Diff, DSP map, and proposed G-DSP map. Meanwhile, a NOKOV motion capture system is used to track markers on the surface of the moving object $\mathcal{O}$ to record its position and velocity as the ground truth. Considering occupied primitives used in our planner, the mean velocity of object along $x$-axis is formulated as $\bar{\mathrm{v}}_x =  \sum_{\mathbf{x}^{(i)} \in \mathcal{O}} \mathrm{v}_x$.
Two cases are validated, i.e. constant-velocity objects and random-velocity objects, where the former is realized using a conveyor belt and the stochastic velocity are induced through random perturbations by human partners.
For quantitative metrics, root mean square error (RMSE) and the standard deviation are used to compare.

For binary classifications, we also set up a typical environment for comparing occupancy performances, see Fig. \ref{fig:MapEnv} (b). To access exact geometries and poses of obstacles, the experiment is conducted in $\rm{PyBullet}$ simulator. Several complex-shaped objects are placed, like bottle, cabinet, multi-parcel and Stanford Bunny model, making it challenging for perception. A fixed RGB-D camera observes the surroundings, in which the robot moves along the given trajectory. For baselines, we additionally remove points on the surface of the robot using a ROS package robot\_body\_filter\footnote{The package is available at https://wiki.ros.org/robot\_body\_filter.} before streaming to maps. 
We generate the ground truth by sampling points from predefined obstacle meshes densely and uniformly, then extract occupied voxels using a realtime Euclidean distance field as \cite{chen2024continuous} did. Moreover, we also utilize raycast to label visibility of voxels. For a concise comparison, we adopt two indicators: F1-score and the area under precision–recall curve (AUC). Three voxel resolutions are studied, including $l = 0.02$ m, $0.03$ m and $0.05$ m. For three particle-based maps, $L_{\max}$ is set to the same, and its value is set to $2.10 \times 10^{6}$, $1.60 \times 10^{6}$ and $0.83 \times 10^{6}$ for three voxel resolutions respectively. The angle resolution $\theta$ is set to $3\degree$ in DSP map and G-DSP map.

\begin{figure}
	\centering
	\setlength{\abovecaptionskip}{0cm}
	\includegraphics[width=1.0\linewidth]{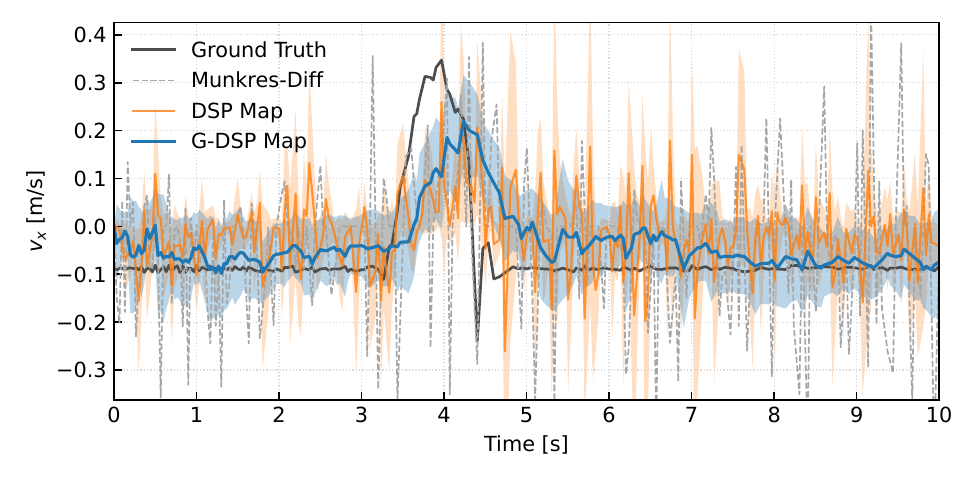}
	\caption{The real-world velocity estimation curves of the object on conveyor belt. The orange and blue shadows indicate the standard deviation of estimation results from DSP map and G-DSP map, respectively. From about $t = 3.4$s to $t = 4.6$s, the parcel is pulled by human partner. }
	\label{fig:VelEst}
\end{figure}

The velocity estimation curves of different maps are shown in Fig. \ref{fig:VelEst} and Table \ref{Table:VelEst}. Our map yields the consistently smoother velocity estimation and lower RMSE compared to baselines. For constant-velocity object, it is observed that the estimation can be completed within 1 s, while it can be recovered within about 2 s from perturbations. While the DSP map also relies on SMC-PHD filter, its velocity estimates are effected by imprecise particle weight updates and arithmetic averaging rather than particle-weighted voxel velocity, resulting in intermediate performance between Munkres and G-DSP map.

\begin{table}[t]
	\fontsize{7.0}{3}\selectfont
	\begin{center}
		\caption{Velocity Estimation Comparison in Manipulation Tasks With Voxel Size Of 0.05 m.}
		\label{Table:VelEst}
		\begin{tabular}{c | c c | c c}
			\hline
			\textbf{Metrics} &
			\multicolumn{4}{c}{\textbf{Case}}\Tstrut\Bstrut \\ \hline
			\textbf{Method} \Tstrut\Bstrut & \multicolumn{2}{c}{Constant-Velocity} \vline & \multicolumn{2}{c}{Random-Velocity} \\ %
			& RMSE $\downarrow$ & Standard Dev. $\downarrow$ & RMSE $\downarrow$ & Standard Dev. $\downarrow$ \\
			\hline 
			\Tstrut\Bstrut Munkres-Diff & 0.142 & --- & 0.167 & --- \\ 
			\Tstrut\Bstrut DSP map & 0.079 & 0.107 & 0.110 & 0.109 \\
			\Tstrut\Bstrut G-DSP map & \textbf{0.035} & \textbf{0.082} & \textbf{0.075} & \textbf{0.083} \\ \hline
		\end{tabular}
	\end{center}
\end{table}

\begin{figure*}
	\centering
	\setlength{\abovecaptionskip}{0.00cm}
	\includegraphics[width=1.0\textwidth]{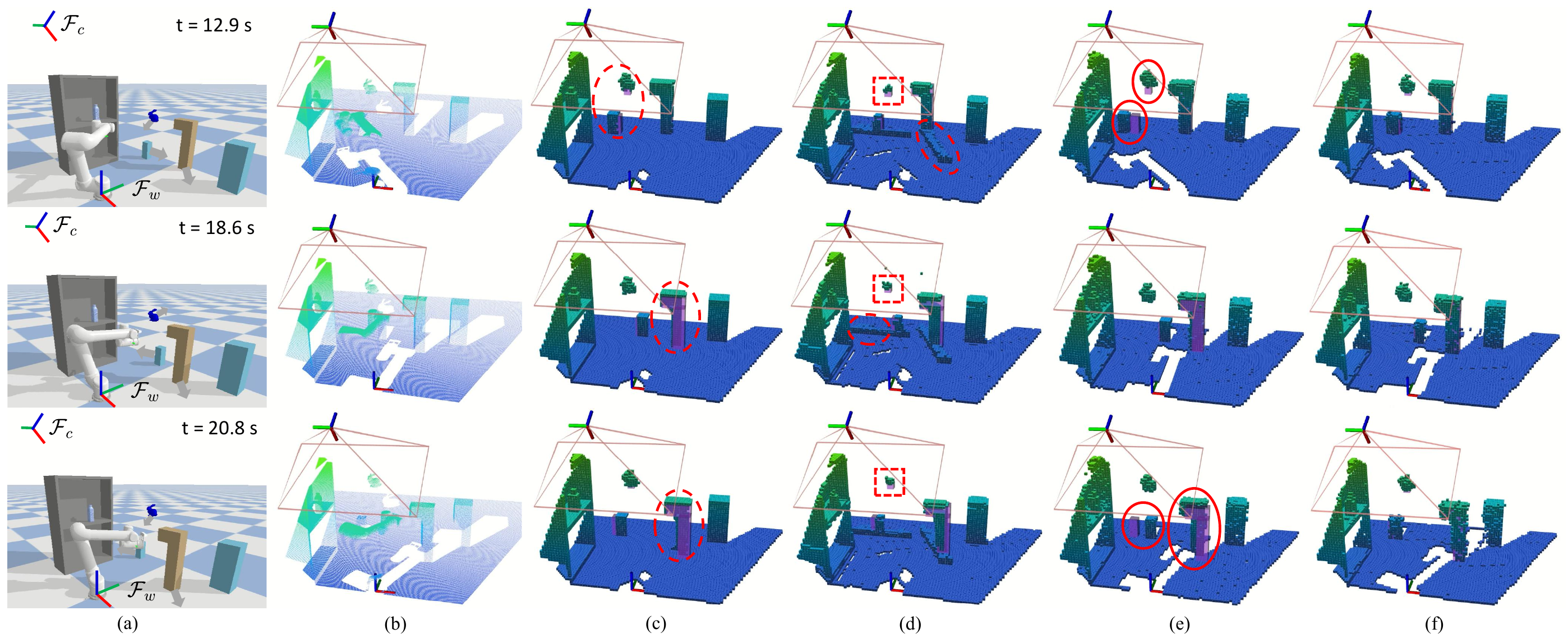}
	\caption{Perception result comparison with a fixed camera when the voxel resolution is $0.02$ m at three time steps. To maintain the dynamic characteristics for manipulation, we only map obstacles in current FOV. The colors of point cloud and voxels indicate their $z$-axis height in the world frame $\mathcal{F}_w$. Meanwhile, the FOV and camera frame $\mathcal{F}_c$ are also visualized. Three translucent purple models represent current real positions of three dynamic objects. The red dashed ellipses show trail noise for probabilistic inertia, while red dashed boxes suggest small objects missing of K3DOM. The red solid ellipses in DSP map show lags resulting from relatively long map latency.  (a) Simulation scenario. (b) Raw point cloud. (c) Ewok \cite{usenko2017real}. (d) K3DOM \cite{min2021kernel}. (e) DSP map \cite{chen2024continuous}. (f) Ours.
	}
	\label{fig:OccupancyResult}
\end{figure*}

See Table \ref{Table:OccEst} for occupancy result of different maps observed in arm manipulation tasks. The proposed G-DSP map achieves the highest average AUC value across all cases, suggesting the most satisfactory overall performance in occupancy estimation. As shown in Table \ref{Table:OccEst}, DSP map demonstrates the comparable performance to the proposed G-DSP map at voxel resolutions of $0.03$ and $0.05$ m. 
When the voxel size is reduced to $0.02$ m, however, Ewok outperforms DSP map, which is attributed to highly increased latency caused by CPU-based fine-resolution mapping. The metrics of K3DOM map are less favorable in manipulation tasks, as it requires an additional classification for static, dynamic and free voxels prior to the particle-based modeling for dynamic ones. Fig. \ref{fig:OccupancyResult} shows snapshots of maps at three time steps $t = \{12.9, 18.6, 20.8\}$ s when voxel resolution is set to $l = 0.02$ m. All three maps are generated with respect to the world frame $\mathcal{F}_w$. Intuitively, both G-DSP and DSP map exhibit dynamic characteristics. In contrast, the baseline Ewok and K3DOM map suffer from trail noise inherent in raycast-based or counting-sensor models, which becomes particularly evident for fast-moving obstacles and leads to false positive occupied voxels (see the red ellipses in Fig. \ref{fig:OccupancyResult}).

\begin{table}[t]
	\fontsize{7.0}{3}\selectfont
	\begin{center}
		\caption{Occupancy Estimation Comparison in Manipulation Tasks. Averaged AUC and best F1 score are reported.
		}
		\label{Table:OccEst}
		\begin{tabular}{c | c c | c c | c c}
			\hline
			\textbf{Metrics} &
			\multicolumn{6}{c}{\textbf{Voxel Size} [m]}\Tstrut\Bstrut \\ \hline
			\textbf{Map} \Tstrut\Bstrut & \multicolumn{2}{c}{0.02} \vline & \multicolumn{2}{c}{0.03} \vline & \multicolumn{2}{c}{0.05} \\ %
			& AUC $\uparrow$ & F1 $\uparrow$ & AUC $\uparrow$ & F1 $\uparrow$ & AUC $\uparrow$ & F1 $\uparrow$ \\
			\hline 
			\Tstrut\Bstrut Ewok map \cite{usenko2017real} & 0.845 & 0.915 & 0.848 & 0.918 & 0.894 & 0.943 \\ 
			\Tstrut\Bstrut K3DOM map \cite{min2021kernel} & 0.242 & 0.357 & 0.307 & 0.426 & 0.411 & 0.550 \\
			\Tstrut\Bstrut DSP map \cite{chen2024continuous} & 0.757 & 0.882 & 0.909 & 0.927 & 0.913 & 0.924 \\
			\Tstrut\Bstrut G-DSP map (Ours) & \textbf{0.910} & \textbf{0.924} & \textbf{0.943} & \textbf{0.958} & \textbf{0.951} & \textbf{0.964} \\
			\hline
		\end{tabular}
	\end{center}
\end{table}

\begin{figure}
	\centering
	\setlength{\abovecaptionskip}{0cm}
	\includegraphics[width=1.0\linewidth]{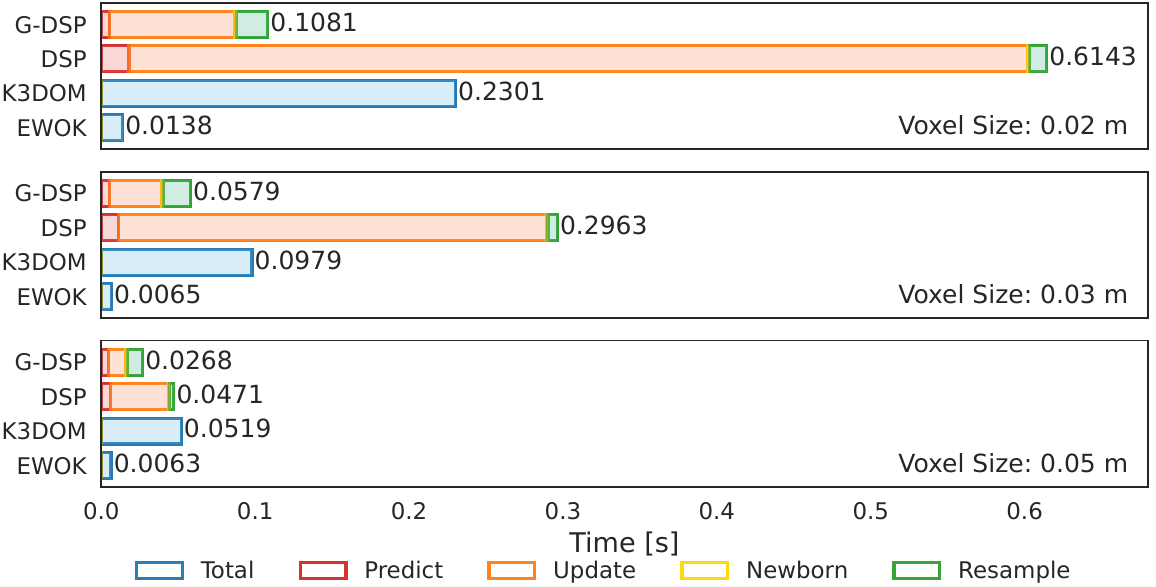}
	\caption{Computation time for four perception methods using different voxel sizes in dynamic environment of Fig. \ref{fig:MapEnv} (b). The averaged step time is reported. }
	\label{fig:MappingTime}
\end{figure}

The computation time for G-DSP map and three baselines is benchmarked in a simulated environment of Fig. \ref{fig:MapEnv} (b) with varying voxel sizes, see Fig. \ref{fig:MappingTime}. As expected, the proposed G-DSP map achieves a significantly lower total step latency than other SOTA particle-based maps. Both prediction and update are substantially faster than those of the DSP map in manipulation tasks. Due to resampling across all occupied voxels in maps, our method incurs a slightly increased resampling cost. While Ewok map attains the shortest mapping time, its static nature and the lack of velocity estimation makes it not suitable in dynamic scenarios. While the total perception time of both G-DSP map and DSP map remain below 100 ms with typically adopted voxel size of 0.05 m in reactive control, the approximated weight update mechanism of DSP map makes it intractable in multi-view setups to mitigate occlusion.

Overall, the experimental results demonstrate that our map consistently outperforms baselines, achieving the overall trade-off between occupancy, velocity estimation and efficiency for dynamic perception in arm manipulations.

\subsection{D-STORM Planning Comparison}
\label{exp2}

\begin{figure}
	\centering
	\setlength{\abovecaptionskip}{0cm}
	\includegraphics[width=1.0\linewidth]{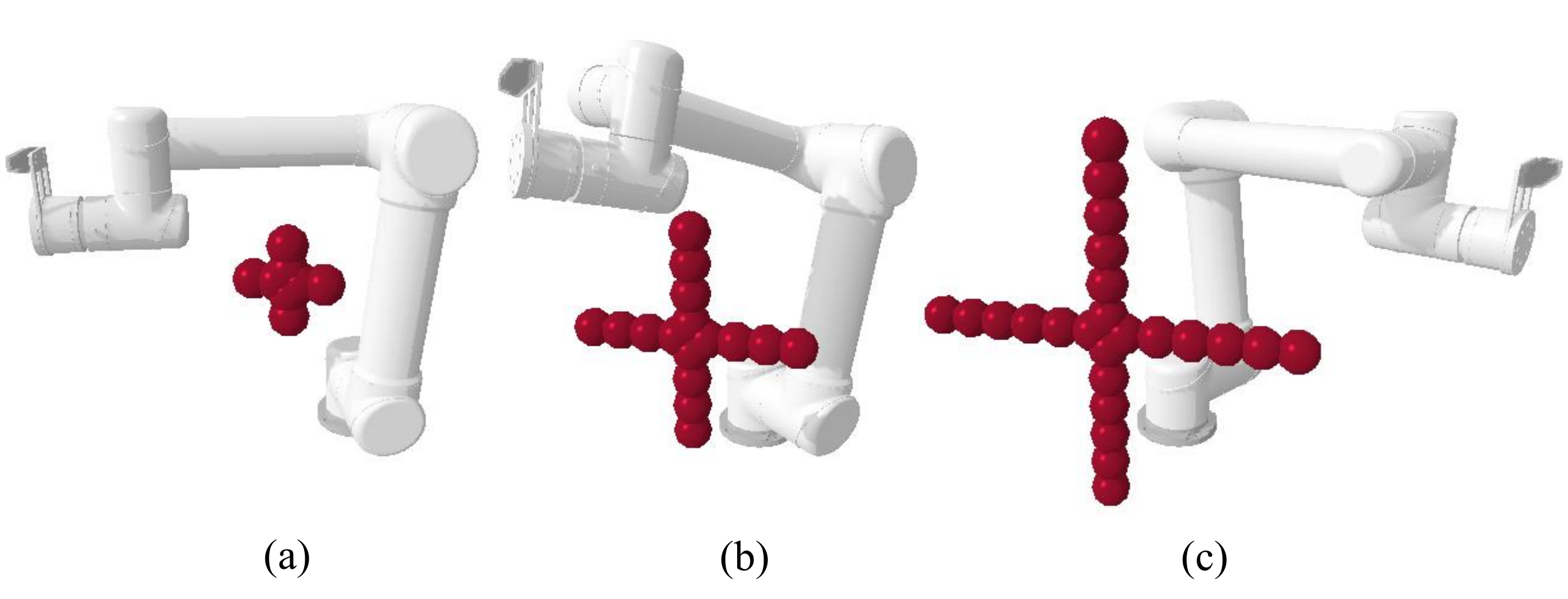}
	\caption{The planning benchmark in dynamic obstacle avoidance. Two-, four-, and six-spheres size cross-shaped obstacle in comparison. Start, intermediate, and goal joint configuration of robot from left to right in goal reaching task. Different obstacle velocities (from 0.0 to 0.2 m/s) are set to evaluate reactive planning performance. }
	\label{fig:PlanningBenchmark}
\end{figure}

To verify hypothesis 2 in Sec. \ref{SubSec:ExpSetup}, we simplify a scenario to a dynamic but known environment. For our benchmark, we refer to the cross-shaped structure placed in front of the robot in \cite{koptev2024reactive} and further improve them to dynamic obstacle cases. We consider a general obstacle dynamics, $\mathbf{p} = \mathbf{p}_0 + \delta \mathbf{p} \sin (\frac{2\pi}{T_o} t)$, and vary the maximum norm of velocity $\mathbf{v}_\mathrm{m} = \Vert \delta \mathbf{p} \frac{2\pi}{T_o} \Vert_2$ from 0.0 to 0.2 m/s to evaluate reactive planning performance. Three obstacle sizes are investigated, see Fig. \ref{fig:PlanningBenchmark}. For baselines, the standard STORM \cite{bhardwaj2022storm} and the DSM-based approach DS-MPPI \cite{koptev2024reactive} are compared. We use the same hyper-parameters as their released implementations. For each method, we conduct 100 goal reaching trials for each case and collect four metrics, including the success rate, the round-trip execution time (i.e. the time from the initial joint configuration to the target and back), the joint-space path length, and the controller's frequency. For D-STORM, the position covariance and velocity covariance of obstacles are set to empirical values, $10^{-3} I_{3 \times 3}$ and $10^{-4} I_{3 \times 3}$, respectively. 
For a fair comparison, we utilize the same simulated UR5 robot and Bernstein polynomial model for three planning methods. Considering different perceptions used, we update the obstacle positions and size for STORM and DS-MPPI at 30 Hz, while the update rate of obstacle information (position, velocity, covariance and size) for D-STORM is set to 10 Hz (close to the worst case of G-DSP map). The number of rollout sequences $K$ for STORM and D-STORM are set to 100, while the number of DS-MPPI is kept at 40.

\begin{table*}[t]
	\fontsize{6.56}{8}\selectfont
	\begin{center}
		\caption{Planning Method comparison of three methods in known static and dynamic environments. Mean (Standard Deviation)}
		\label{Table:PlanningEst}
		\begin{tabular}{c c | c c c | c c c | c c c | c c c}
			\hline 
			\textbf{Obs.} & \textbf{Obs.} & \multicolumn{3}{c}{\textbf{Success Rate} $\uparrow$} \vline & \multicolumn{3}{c}{\textbf{Round-Trip Time}  [s]} \vline & \multicolumn{3}{c}{\textbf{Path Length} [rad]} \vline & \multicolumn{3}{c}{\textbf{MPC Frequency} [Hz] $\uparrow$} \Tstrut\Bstrut \\ 
			\textbf{Size} & \textbf{Max Vel} & STORM & DS-MPPI & Ours & STORM & DS-MPPI & Ours & STORM & DS-MPPI & Ours & STORM & DS-MPPI & Ours \\
			\hline
			\Tstrut\Bstrut & 0.0 m/s & 1.00 & 1.00 & 1.00 & 10.59 (0.86) & 4.97 (0.17) & 11.75 (1.02) & 10.75 (0.99) & 11.66 (0.26) & 12.16 (0.50) & 44.0 & 9.9 & 43.8 \\
			2 \Tstrut\Bstrut & 0.1 m/s & \textbf{1.00} & 0.99 & \textbf{1.00} & 11.49 (2.02) & 6.25 (1.05) & 11.85 (1.11) & 12.43 (1.18) & 13.86 (2.05) & 12.17 (0.63) & 43.1 & 9.9 & 42.4 \\
			\Tstrut\Bstrut & 0.2 m/s & 0.88 & 0.96 & \textbf{0.99} & 11.53 (2.00) & 5.67 (1.13) & 12.45 (2.01) & 12.40 (1.50) & 12.39 (1.60) & 12.32 (0.96) & 40.5 & 9.8 & 41.0 \\ \hline
			\Tstrut\Bstrut & 0.0 m/s & 1.00 & 1.00 & 1.00 & 12.76 (1.21) & 5.31 (0.24) & 14.28 (1.11) & 12.25 (0.30) & 11.77 (0.60) & 12.44 (0.27) & 29.4 & 9.7 & 29.8 \\
			4 \Tstrut\Bstrut & 0.1 m/s & 0.93 & \textbf{1.00} & \textbf{1.00} & 13.50 (1.78) & 5.54 (1.74) & 14.89 (2.10) & 12.89 (1.26) & 11.84 (0.93) & 12.80 (0.97) & 28.8 & 9.9 & 29.8 \\
			\Tstrut\Bstrut & 0.2 m/s & 0.83 & 0.92 & \textbf{0.97} & 13.20 (2.02) & 6.18 (2.37) & 16.74 (3.17) & 12.67 (1.21) & 12.79 (1.27) & 14.43 (2.23) & 28.3 & 9.9 & 29.0 \\ \hline
			\Tstrut\Bstrut & 0.0 m/s & 1.00 & 1.00 & 1.00 & 14.62 (1.68) & 6.53 (0.70) & 15.92 (1.04) & 13.36 (0.72) & 13.70 (1.33) & 13.13 (0.33) & 25.6 & 9.8 & 25.4 \\
			6 \Tstrut\Bstrut & 0.1 m/s & 0.88 & 0.97 & \textbf{0.98} & 16.27 (4.63) & 6.94 (1.47) & 17.46 (3.04) & 14.34 (3.37) & 14.35 (1.80) & 13.93 (1.77) & 25.6 & 10.0 & 25.3 \\
			\Tstrut\Bstrut & 0.2 m/s & 0.73 & 0.87 & \textbf{0.96} & 13.63 (2.35) & 5.95 (1.75) & 17.97 (3.23) & 12.39 (0.85) & 12.72 (1.05) & 14.85 (2.09) & 24.8 & 10.1 & 25.6 \\
			\hline
		\end{tabular}
	\end{center}
\end{table*}

Table \ref{Table:PlanningEst} summarizes simulation results of collision avoidance with ideal obstacle positions and velocities. In dynamic environments, the proposed D-STORM achieves the substantially higher success rate compared to two baselines. While our approach incurs longer round-trip time and trajectory length, it proactively predicts obstacle motion in planning, mitigating failures caused by rapid environmental changes or convergence to the local minima. In contrast, DS-MPPI exhibits the shortest execution time and shows reactive collision avoidance under static and slowly moving obstacles due to its high-frequency dynamical system. However, when obstacle velocities increase to 0.2 m/s, DS-MPPI becomes constrained by limited parallelism and modulation rate (about 10 Hz on GPU), which leads to reduced performance under fast obstacle motions. 

As can been seen, the update frequency of STORM and D-STORM are sensitive to the number of obstacles. Nonetheless, we restrict the maximum considered obstacles to 20, resulting in the minimum frequency of about 25 Hz. Moreover, planning in the joint acceleration space can result in smoother motions for both methods. Hence, by explicitly considering the obstacle dynamics and incorporating predictive capability, our method demonstrates enhanced safety in dynamic environments.

\subsection{Simulation Pipeline Comparison and Discussion}
\label{exp3}

\begin{figure}
	\centering
	\setlength{\abovecaptionskip}{0cm}
	\includegraphics[width=0.98\linewidth]{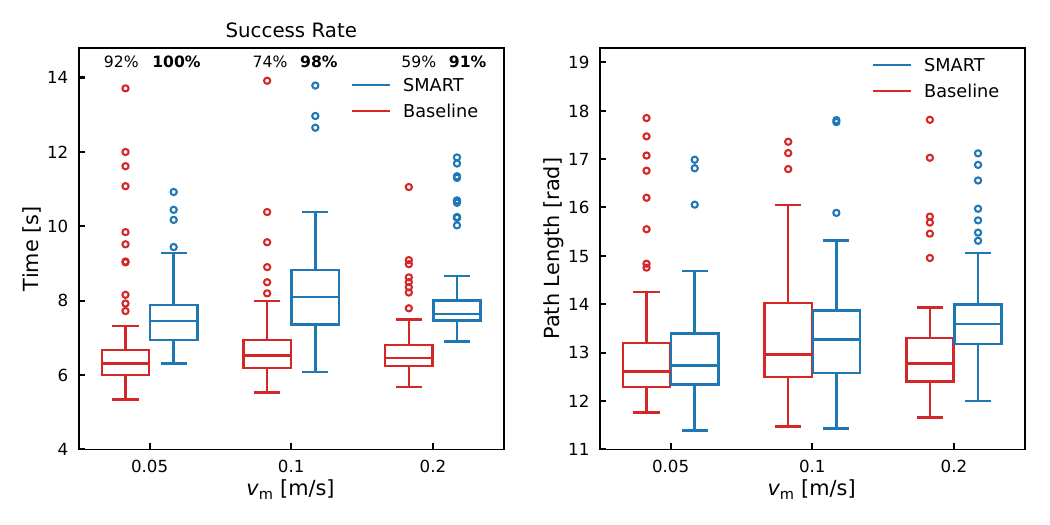}
	\caption{Comparison of success rate, trajectory execution time and path length for sensor-based planning in simulated unstructured dynamic environments. Cases with time exceeding 14 s or path length exceeding 18 rad are clipped. }
	\label{fig:SensorBasedPlanning}
\end{figure}

\begin{figure*}
	\centering
	\setlength{\abovecaptionskip}{-0.06cm}
	\includegraphics[width=1.0\textwidth]{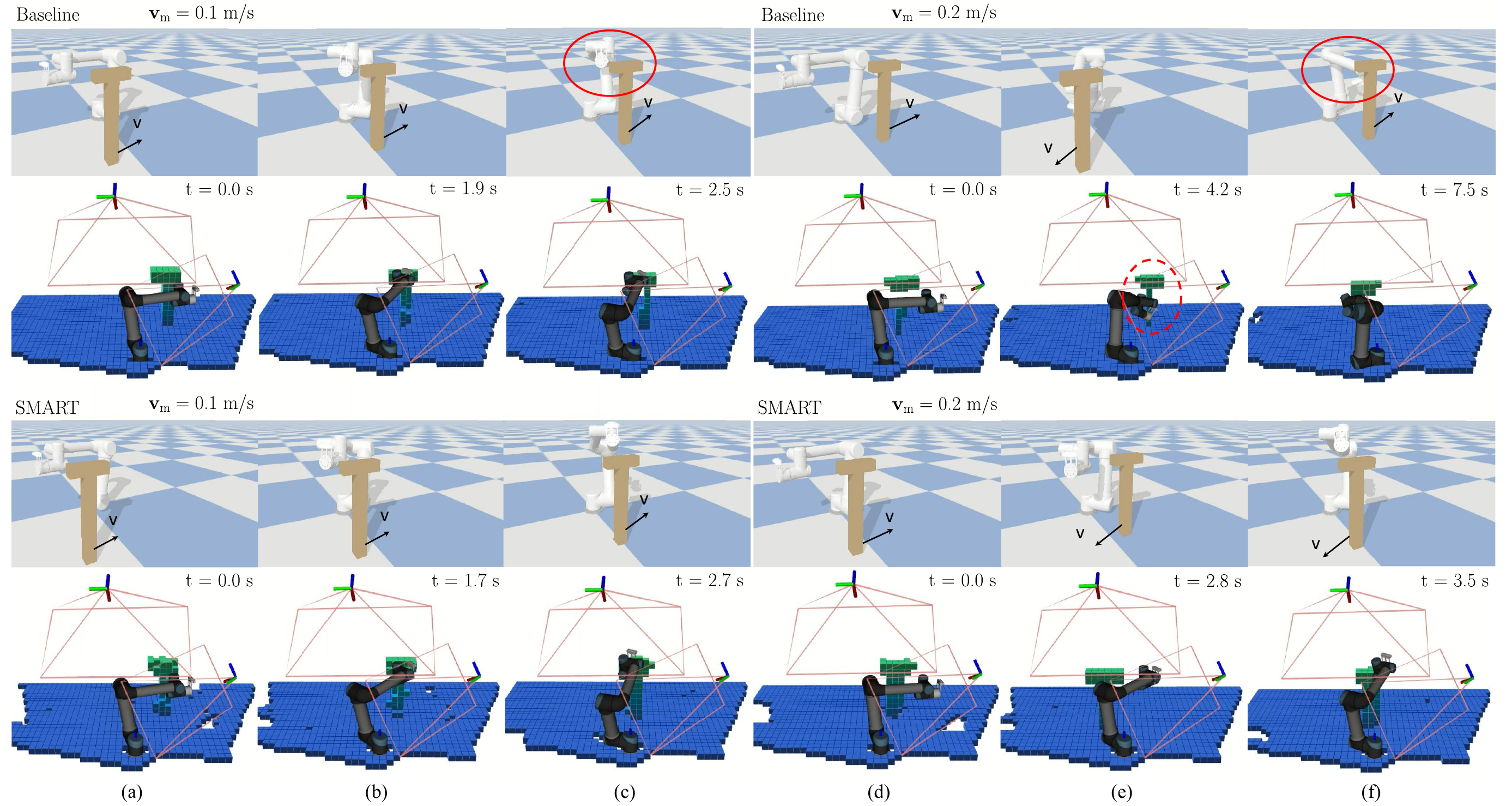}
	\caption{Sensor-based planning comparison with two cameras when the maximum velocity is $0.1$ m/s and $0.2$ m/s in scenario 1. The top two lines show the planning procedure of baseline, while the bottom two lines indicate snapshots of ours. (a)-(c): $\mathbf{v}_\mathrm{m} = 0.1$ m/s. Although the baseline framework perceives the moving obstacle and generate an avoidance trajectory, it fails to anticipate the obstacle’s motion and consequently results in collision. (d)-(f): $\mathbf{v}_\mathrm{m} = 0.2$ m/s. The raycast-based map suffers from probabilistic inertia for fast-moving obstacles, making robot failed to perceive the approach of the T-shaped obstacle in time (see red dashed ellipses of Line 2). By explicitly incorporating dynamic characteristics, SMART consistently achieves collision-free motion.
	}
	\label{fig:SensorBasedPlanningResult}
\end{figure*}

\begin{figure*}
	\centering
	\setlength{\abovecaptionskip}{-0.06cm}
	\includegraphics[width=1.0\textwidth]{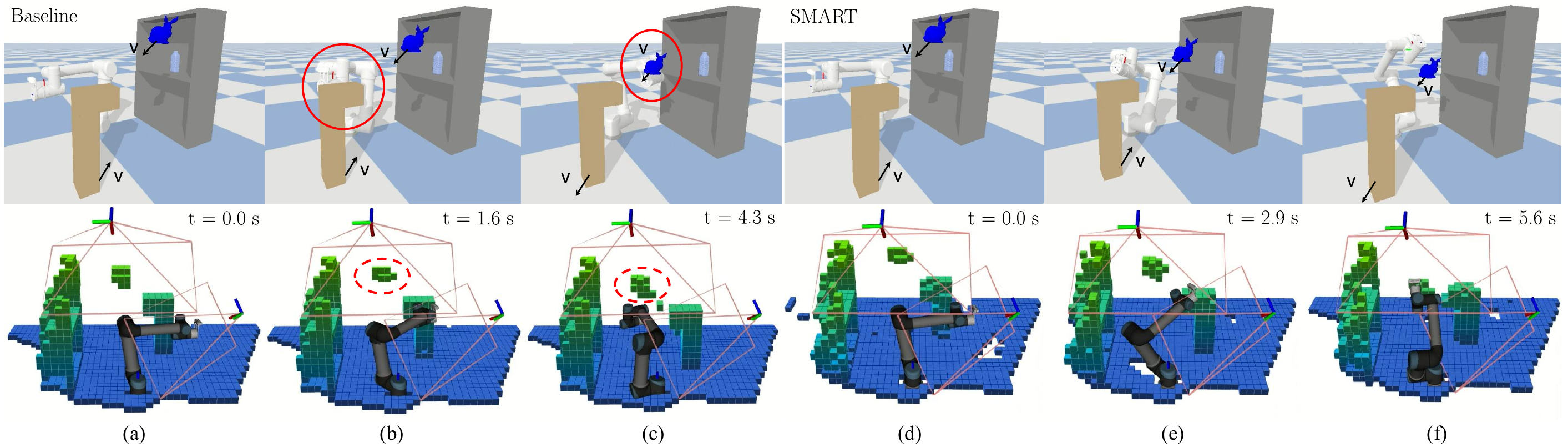}
	\caption{Sensor-based planning in scenario 2 with multiple obstacles. (a)-(c): Under baseline method, the robot could not respond sufficiently fast to moving multi-parcel, leading to near-collision contacts, and collides with a dynamic bunny model approaching from above (red solid ellipses of Line 1). Red dashed ellipses indicate delayed mapping using raycast-based perception. (d)-(f): In contrast, SMART consistently successfully avoids both dynamic obstacles.
	}
	\label{fig:SensorBasedPlanningMultiObjectResult}
\end{figure*}

In this section, we compare the performance of the proposed perception-planning framework SMART with the conventional and commonly used pipeline. Two dynamic simulation scenarios are constructed using a UR5 manipulator and two fixed RealSense cameras (see Fig. \ref{fig:SensorBasedPlanningResult}-\ref{fig:SensorBasedPlanningMultiObjectResult}). In particular, we compare against the STORM planner \cite{bhardwaj2022storm} with dual-view raycast-based perception and implement it using the same setup. To evaluate their collision-free motion generation performance to varying obstacle velocities, we consider a T-shaped dynamic obstacle, whose dynamics follows sinusoidal curve defined in Sec. \ref{exp2}, with the maximum velocities set to 0.05, 0.10, and 0.20 m/s. The manipulator moves from an initial configuration to a target one while avoiding the moving obstacle. For both the baseline and SMART, 100 independent cases are conducted under each velocity. Trajectory execution time, path length and success or failure are recorded, from which the success rate and metrics under successful trials are analyzed. Considering whole-body planning, voxel sizes for both maps are set to 0.05m.

We plot the results in Fig. \ref{fig:SensorBasedPlanning}. While the baseline (STORM + Ewok) shows relatively high success rate and reach slightly lower time under low-speed obstacles, it degrades significantly under fast-moving obstacles. As can be seen, the metrics in dynamic scenarios naturally exhibit the long-tailed distribution due to stochastic obstacle motion and multimodal robot policy (e.g. avoiding the obstacle from above or below). For path length of reactive planning, SMART achieves a tighter interquartile range compared to the baseline, indicating more consistent behavior in collision avoidance. As expected, by explicitly considering robot-obstacle dynamics, the proposed framework achieves significantly higher success rate, slightly longer execution time and trajectories with enhanced safety. In contrast, the more aggressive behavior of the baseline, while efficient in simulation, poses notable safety concerns in real-world deployment, where the success rates drop to 74\% and 59\% respectively with moderate and fast obstacles (0.1 and 0.2 m/s). The snapshots of two collision-free motion generation cases are showed in Fig. \ref{fig:SensorBasedPlanningResult}.

To qualitatively compare the sensor-based planning performance under multiple obstacles, we construct a more complex simulation environment (scenario 2) comprising two dynamic obstacles (a multi-parcel and a bunny model) together with several static obstacles (see Fig. \ref{fig:SensorBasedPlanningMultiObjectResult}). Using the baseline, the robot struggles to consecutively avoid two dynamic obstacles, leading to frequent near-collision or failure cases. Owing to the dynamic characteristics of SMART, the robot can successfully avoid both moving obstacles and reach the target configuration in uncertain cases. The results highlight the advantage of the proposed framework in handling multiple dynamic obstacles.

From the quantitative to qualitative results, SMART demonstrates superior reactive performance for perception and collision avoidance of multiple slow and moderate obstacles with complex shapes.

\subsection{Real-World Applications on Safe Planning}
\label{exp4}

\begin{figure*}
	\centering
	\setlength{\abovecaptionskip}{-0.02cm}
	\includegraphics[width=1.0\textwidth]{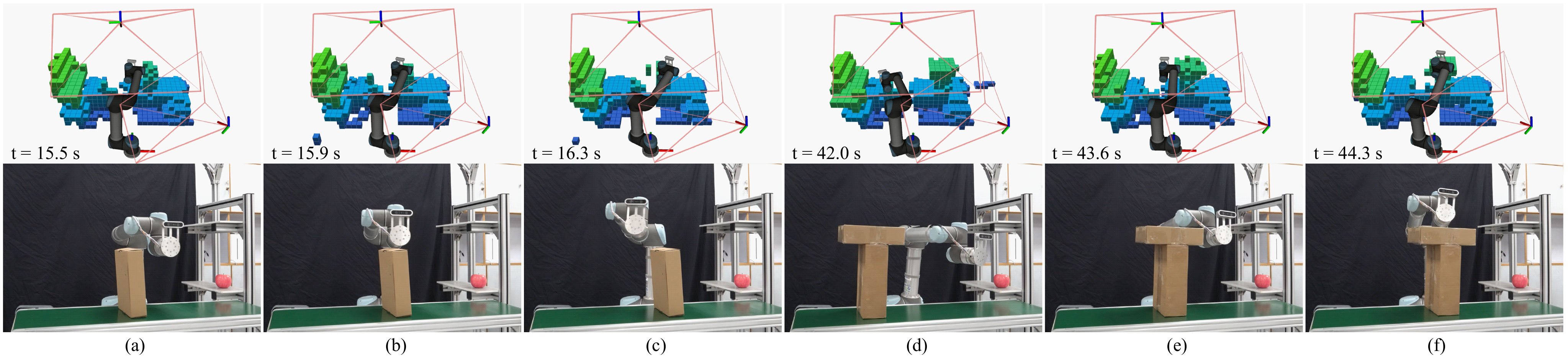}
	\caption{Sensor-based planning in real-world tasks. The six-DOF robot reactively avoids parcels that are transported on conveyor belt using SMART. (a)-(c): The robot avoids small parcel and continues to the target position. (d)-(f): More complex-shaped multi-parcel moves at 0.1 m/s, which is perceived by sensors and avoided by robot. The onboard camera can be used for grasping when reaching target positions in future applications.
	}
	\label{fig:RealworldExp}
\end{figure*}

Real-world experiments are performed using a 6-DOF UR5 robot on a fixed base (see Fig. \ref{fig:RealworldExp}). We consider the scenario involving multiple static and dynamic obstacles, where several parcels are transported on the conveyor belt at the speed of 0.1 m/s. The robot manipulator alternately executes reaching tasks between two robot joint configurations while avoiding collisions with surroundings. Realtime perception is performed using G-DSP map with the extent of $[2.2 \times 2.2 \times 1.5] \mathrm{m}$, and the D-STORM planner is utilized to demonstrate reactive collision avoidance performance. Two fixed RealSense depth cameras are deployed to stream synchronized point clouds, with one mounted laterally and the other positioned behind and above the robot. Then the robot is controlled at 125 Hz by a low-level controller.

In real-world deployments, two important aspects merit further discussion. First, real depth cameras exhibit significantly higher measurement noise than idealized observations assumed in simulation. Despite filtering for point cloud, particle-based perception 
remains subject to estimation errors, which leads to spurious occupancy near obstacles. Such perception noise will induce oscillatory behaviors for well-designed planner. Hence, we employ softmax-based sum over obstacle primitives rather than maximum operator in real-world deployments, 
\begin{equation}
	c_{col} (\mathbf{q}_h) = \frac{1}{\alpha_o} \log \sum_{i = 1}^{N_o^c} \exp (\alpha_o c_{col}(\mathbf{x}^{(i)}_h, \mathbf{q}_h)),
	\label{softmax_cost}
\end{equation}
which provides a smooth and risk-sensitive approximation of the maximum cost while remaining robust to perception noise in real-world deployment, $\alpha_o$ is set to 25.0 in experiments.

On the other hand, latency and computational load should also be considered in deployments. In real-world settings, the perception-planning pipeline will inevitably introduce the non-negligible latency. In our framework, tensor-based method are employed to accelerate both perception and reactive planning. However, running these two modules concurrently on a GPU could lead to implicit serialization of workloads and reduced throughput. We evaluate the proposed sensor-based planning framework using a single GPU to provide preliminary evaluation of its performance and to highlight advantages in dynamic environments. In practical deployments, perception and control can be executed on separate GPUs, thereby avoiding resource contention and alleviating performance degradation.

\begin{figure*}
	\centering
	\setlength{\abovecaptionskip}{-0.02cm}
	\includegraphics[width=1.0\textwidth]{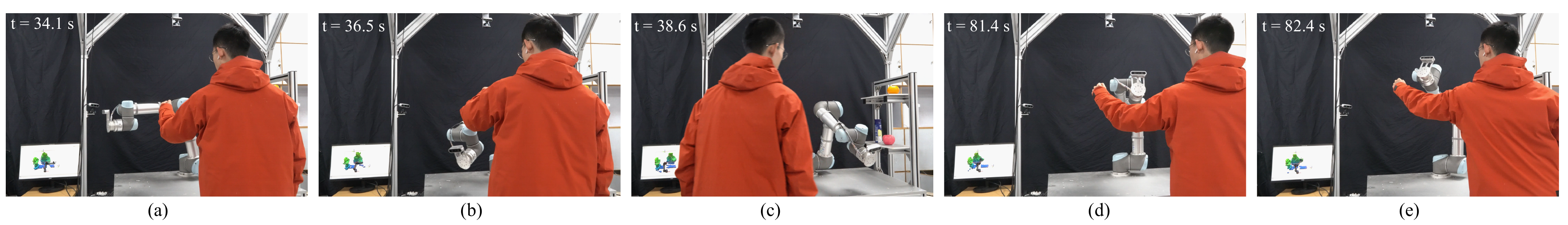}
	\caption{The experiment of pHRI. A six-DOF robot reactively avoids the human using SMART in the collaborative workspace. Robot avoids the raised arm to reach the target configuration.
	}
	\label{fig:RealworldHRI}
\end{figure*}

In the unstructured and dynamic scenario, the arm successfully perceives and avoids two moving parcels and other static obstacles, as illustrated in Fig. \ref{fig:RealworldExp}.  Using SMART, the robot effectively avoids parcels from above. In contrast, the baseline framework does not explicitly consider dynamic objects and is not applicable to safe obstacle avoidance in such environments. Fig. \ref{fig:RealworldHRI} demonstrates the snapshots of pHRI where the human partner dynamically behaves in the workspace.

\subsection{Discussions and Limitations}

In this article, we investigated a sensor-based reactive motion generation method in dynamic and trivially unstructured environments. It is assumed that obstacles are static or geometrically known for conventional planning methods. Our work SMART innovatively introduced particle-based perception into collision-free planning for manipulators, revealing an insight of unified tensor optimization formulation. The particle-based G-DSP map designed for manipulation provided the effective estimation of positions, velocities, and uncertainty of multiple arbitrary-shaped obstacles. In particular, the tight correlation between D-STORM and G-DSP theoretically solved the problem of collision avoidance for dynamic obstacles by propagating robot-obstacle dynamics. In brief, SMART preliminarily considers the unstructured and dynamic environments in real-world manipulations, providing a novel and general framework for collision avoidance in pHRI, autonomous exploration and many factory applications.

Particle-based maps have been initially explored in UAVs or AGVs and have various formulations for occupancy, velocity and uncertainty estimation. Previous works (maps) have either utilized Dirichlet distribution to estimate velocities of voxels, or directly predicted future occupied state via particle dynamics. For high-DOF manipulators, the numerous particles will highly increase computational load of robot-obstacle collision checking, especially for MPPI-based motion generation using GPU. The randomness of particles will also introduce severe noise to planners. Hence, we approximately adopt voxels as obstacle primitives. Different from most particle-based maps that have been open-sourced, we implement G-DSP perception using parallel operations with library $\rm{PyTorch}$, and provided the embedding for kinematics of high-DOF manipulators. The performance of parallel implementation turned out to be about 2x faster than SOTAs in manipulation tasks. By evaluations, particle pool and tensor optimization effectively provide richer perception for manipulators from a model-based perspective.

Nevertheless, the current implementations for collision-free motion generation is limited in \textit{trivially unstructured} scenarios and the contact-free planning. Since only predicting particles outside FOV without subsequent update may lead to inaccurate occupancy estimation, the current implementation focuses on avoiding obstacles that are observed within FOV. Several directions remain to improve sensor-based reactive planning. First, real-world manipulations tends to involve objects of different sizes, motivating the integration of multi-resolution maps to better balance accuracy and efficiency. As particle-based maps suffer from the particle degeneracy outside FOV, incorporating static obstacle identification and memory module is an important extension. Also, fusing visual sensing and tactile feedback can further enrich multimodal perception for manipulators. If there exists some large obstacles of complex shapes, such as the large concave ones that are partially observed, the proposed method may be also stuck in the local minima. In such cases, it is promising to consider active viewpoint generation in our reactive planning framework, balancing the motion optimality and safe geometric exploration. We leave them to future works.

\section{Conclusion}
\label{Sec:Conclusion}

We motivated this work by noting tight connections between the effectiveness of continuous particle-based mapping and the struggle of MPPI-based planning in dynamic environments. By explicitly revealing and utilizing the robot-obstacle dynamics, we have presented a sensor-based planning framework to introduce a tensorized global particle-based perception to dynamic-aware MPPI to enhance the collision avoidance capabilities in unstructured environments. Compared to existing perception-planning pipelines for manipulators which utilize the raycast-based map and vanilla MPPI, we have contributed an insightful alternative in reactive and sensor-based planning.

First, we leverage the SMC-PHD filter and tensor operations to propose a particle-based G-DSP map for manipulators. The velocities and spatial uncertainty of voxels are explicitly considered and efficiently calculated, allowing planners to incorporate obstacle dynamics from model-based perceptive. Next, enabled by richer perception, we presented a novel D-STORM formulation which jointly propagates robot-obstacle dynamics, and adaptively refine optimal trajectory under perception and control uncertainty. Finally, by tightly coupling particle-based map and sampling-based control, SMART bridges perception and reactive planning in dynamic and unstructured scenarios. With extensive theoretical analysis and empirical evaluation, we demonstrate the efficacy of SMART on high-dimensional manipulators operating in intermediately dynamic and unstructured environments, where deterministic assumption often fail. 
We believe that in the future, the highly optimized version of the proposed framework can be used for practice foundation of reactive motion generation in unstructured manipulation and safe pHRI. Future work will include wider FOV perception, the combination with learning-based active exploration, and acceleration-aware obstacle dynamics, toward more intelligent and robust collision-free planning.

\section*{Appendix \\Preliminary Extension to Multi-View Perception}

In method sections, we discuss the core mapping algorithm with an onboard/fixed depth sensor. With multi-view particle-based perception, the collision avoidance performance of the manipulator is better. Suppose there are two cameras mounted on $\mathbf{x}_C^1$ and $\mathbf{x}_C^2$ respectively. Upon receiving two point clouds $Z_t^1$, $Z_t^2$ from two depth sensors, the voxel filtering is firstly applied, transforming both point clouds into the world frame and merging into a point cloud $\tilde{Z}_t$. Subsequently, points on the surface of manipulator are removed using robot SDF geometry. Note that two point cloud partially overlap, we perform voxel filtering again to remove redundancy. The occlusion handling for particles in map is performed in two observation pyramids, as detailed in Alg. \ref{alg5}. Then the remaining steps follow Alg. \ref{alg3}.

\begin{algorithm}[t]
	\label{alg5}
	\caption{Update for Dual-View System}
	\SetAlgoLined
	\DontPrintSemicolon
	\KwIn{Particle Tensor $\tilde{X}_{t} (\tilde{X}_{s, t})$, Point Tensor $\tilde{Z}_t$, \mbox{Camera Positions $\mathbf{x}_C^{1}, \mathbf{x}_C^{2}$, Thickness $\epsilon_H$}}
	\KwOut{Updated Particle Tensor $\tilde{X}_t$}
	
	\For{$i = 1, 2$}{
		$\mathbb{P}_t^{i} \leftarrow \mathbf{AssignParticlesToPyramids}(\tilde{X}_{t}, i)$ \\
		$d_z^i \leftarrow \Vert \tilde{Z}_{t} - \mathbf{x}_C^{i} \Vert_{2}$ \\
		$\mathbb{P}_z^i \leftarrow \mathbf{AssignPointsToPyramids}(\tilde{Z}_{t}, d_z^i)$ \\
		\tcp*[l]{\footnotesize{Max observation distance tensor}}
		$d_{\max}^i \leftarrow \underset{j = 1, ..., M_{\max}/N_p}{\max} \mathbb{P}_z^i[:, j] \in \mathbb{R}^{N_{p}^i}$ \\
		\tcp*[l]{\footnotesize{Max particle distance tensor}}
		\mbox{$d_{\max}^i \leftarrow \mathbf{MaxDistForParticles}(d_{\max}^i, \mathbb{P}_t^i) \in \mathbb{R}^{L_{s, t}}$} \\
		$\tilde{X}_{f, t}$ $\leftarrow \tilde{X}_{f, t} \cup \tilde{X}_{t}[\Vert \tilde{X}_{t} - \mathbf{x}_C^i \Vert_{2} - d_{\max}^i \leq \epsilon_H]$ \\
	}
	$C(\tilde{Z}_{t}) \leftarrow \mathbf{Denominator}(\tilde{Z}_{t}, \tilde{X}_{f, t})$ \qquad\quad $\triangleright$ Eq. \eqref{c_batch} \\
	$\mathbf{UpdateWeights}(\tilde{Z}_{t}, \tilde{X}_{f, t}, \tilde{X}_{t})$  ~~~~~ $\triangleright$ Eq. \eqref{G_batch}, \eqref{update_tensor}
\end{algorithm}

%

\bibliographystyle{IEEEtran}
\bibliography{references}

\newpage
\vspace{11pt}

\newpage
\vfill

\end{document}